\documentclass[10pt,twocolumn,letterpaper]{article}

\usepackage{cvpr}
\usepackage{times}
\usepackage{graphicx}
\usepackage{amsmath}
\usepackage{amssymb}

\usepackage{bm}
\usepackage{bbm}
\usepackage{xcolor}
\usepackage[export]{adjustbox}
\usepackage{multirow,bigstrut}

\def\x{\mathbf x}

\def\Z{\mathcal Z}

\def\Y{\mathcal Y}

\graphicspath{{ims/bird-shot0_raws/}{ims/garden-boy_raws/}{ims/juggler/}{ims/bird-shot00_raws/}{ims/bird-shot0-baseline-1/}{ims/bird-shot0-baseline-2/}{ims/garden-boy_raws/}{ims/garden-baseline-1/}{ims/garden-baseline-2/}{ims/ocean_boat_raws/}{ims/ocean-boat-base-ldb50-nomot/}{ims/ocean-boat-baseline-2/}{ims/}{ims/cat_raws/}{ims/cat_baseline-1-predict/}{ims/cat_baseline-2-predict/}}
 
\cvprfinalcopy 

\def\cvprPaperID{5231} 

\ifcvprfinal\fi
\begin{document}

\title{  Motion Selective Prediction for Video Frame Synthesis}

\author{V\'eronique Prinet\\
The Hebrew University of Jerusalem\\
{\tt\small vprinet@gmail.com}
}

\maketitle

\begin{abstract}
Existing conditional video prediction approaches train a network from large databases and generalise to previously unseen data. We take the opposite stance, and introduce a model that learns from the first frames of a given video  and extends its content and motion, to, \eg, double its length. To this end, we propose a dual network that can use in a flexible way both dynamic and static convolutional motion kernels, to predict future frames. The construct of our model gives us the the means to efficiently analyse its functioning and interpret its output. We demonstrate  experimentally the robustness of our approach  on  challenging videos in-the-wild and show that it is competitive \wrt related baselines. 
\end{abstract}



\section{Introduction}

We consider  the problem of motion prediction for future frame synthesis. While the vast majority of the recent lite\-rature in the field is dedicated to learning forecasting models from (relatively) large databases, we focus our attention on learning from few samples.  
Being able to learn efficiently from small data,  exploiting a {\em good motion representation},  opens the door to a variety of new applications. 

We explore for the first time predictive mo\-dels that are {\em domain-agnostic}  but  {\em data-specific}. Our aim is to learn a model  of a dynamic scene in the wild from a single video clip, and to extend/extrapolate its content and motion to, \eg, double its length. We are interested in any natural motions, such as,  a  bird in flight (see figure~\ref{fig:demo}) or the gesture of a juggler (see the Section~\ref{sec:res}).

	\bgroup
    \def\arraystretch{0}
	\begin{figure}[t]
	\begin{center}
	\begin{tabular}{cc}
	\multirow{4}{*}[3.65cm]{ \includegraphics[width=0.45\linewidth, height=15.3cm]{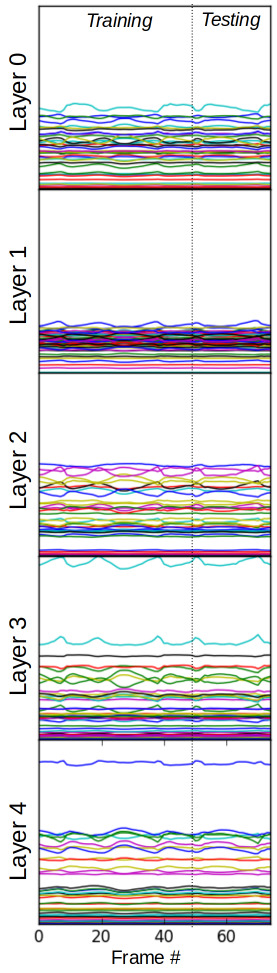}} & 
  		\includegraphics[width=0.43\linewidth, cfbox=blue 1pt 0pt]{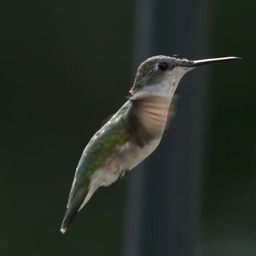} \\ 
    		 &\includegraphics[width=0.43\linewidth, cfbox=green 1pt 0pt]{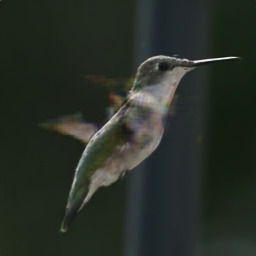}\\
    		 &\includegraphics[width=0.43\linewidth, cfbox=green 1pt 0pt]{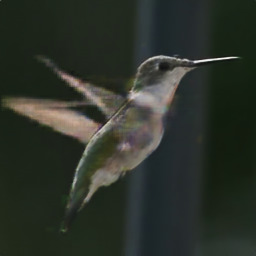}\\
    		 &\includegraphics[width=0.43\linewidth, cfbox=green 1pt 0pt]{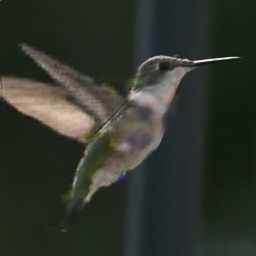}
   	\end{tabular}
	\end{center}
	.\vskip0mm
   	\caption{ Conditioned on a few context frames (blue frame), the transformer, $G()$, generates future ones (green frames). The selector  modulates dynamically the amplitude of $G()$'s motion kernels.}
	\label{fig:demo}
	\end{figure}
	\egroup

Learning a predictive model from a single video  in the wild is challenging: 1) the generic nature of the natural motion we are seeking to model is not suitable for  loss-specific or architecture-specific networks of most existing methods; 2) the choice of videos-in-the-wild implies a model capable of robust background-foreground decomposition,  to be able to recover large background regions occluded by the foreground in previous frames  --something that no work of our knowledge so far has demonstrated; 3) learning from a short clip requires a quick and efficient convergence of the model at training time.

Our model is related to two different lines of work tackling the issue of motion prediction.  The first one is concerned with  dynamic filters, \ie methods that infer input-dependent weights of a convolutional or LSTM network at each time-step, and apply these filters to a frame  to predict the next one~(\eg~\cite{fin:16,bra:16}). The second one refers to  disentangled representations from unsupervised learning~(\ie~separating the causes from the effect of an action) --approaches that usually implicitly assume  simple background or semi-rigid motion~\cite{den:17, hsi:18, vil:17a}. 

Different from those prior works, we let our model jointly use  dynamic and static elementary convolutional kernels at multi-scales. It learns in a unsupervised manner how to associate static kernels to the generation of the  background image and dynamic ones to the generation of the moving  foreground. Our motion representation is based on a dual network: one that learns kernels, and a second one which dynamically selects the best subset for  next frame prediction. Inspired initially by the mechanism of Direction Selective (DS) cells in the retina~(see \cite{bri:11,sun:06}), it is extremely simple and does not require a tailored loss or net architecture.

An other body of work related to ours is  learning from few data~\cite{wei:02,rus:18}. This domain is predominantly covered nowadays by the literature in meta-learning (aka, one-shot-learning). 
While the setting of metalearning is not ours (it relies on a large database to learn a meta-network), some of the findings are related: in particular,  good initialization, or in our case, rapid exploration of distinct optimal solutions at at an early stage of the training, is key to an efficient convergence of the net.

Our contributions can be summarized as follows: 
\begin{enumerate}
    \item We introduce a deep motion model for video frame prediction. To our knowledge, this is the first work which investigates learning from a single video-clip for {\em domain-agnostic}  but  {\em data-specific} application.
    \item We show that it is possible to analyse {\em how} the model operates to generate the future and what it has learnt from the data.   
    \item We validate our approach on natural videos with cluttered background, occlusion, multiple and complex motions, and no particular semantic domain, with mid-range\footnote{We use the term `mid/long range' rather than `long-term', because the difficulty of the task does not lie so much on the number of frames to generate, but rather on the amplitude of the motion between the first and the last frame.} (10-30 frames) prediction. To our knowledge, no previous work have demonstrated results on such challenging data. Our video sequences  data will be made publicly available. 
\end{enumerate}

\section{Related Work}

\paragraph{Motion representation}

Motion representation is a long-standing open problem in visual perception studies and computer vision~\cite{chan:09, luc:81,bla:96,fle:00, wei:02}. Visual illusions show strong evidence that the perceived motion between consecutive images strongly depends on the image structure itself~\cite{wei:02}. The first attempt to develop a parametric statistical model that explicitly captures  the conditional dependence between the flow field and the input image structure, might be attributed to Sun \etal.~\cite{sun:08}. More recently, variational auto-encoders (VAE) have been shown to be an efficient means to modelling motion with learned prior~\cite{den:18}. Our model also learns input-dependant constraints on the flow field, albeit in a non-stochastic manner.

\paragraph{Video texture synthesis} Dynamic texture (or textured motion) are sequences of images of moving scenes or objects that exhibit certain harmonic or stationary properties in time, often encountered in natural scenes (\eg, fluid flow, clouds). Early parametric~\cite{soa:01, zhu:03}  and non-parametric~\cite{bar:01} approaches were mostly suitable to model global dynamic systems. Layered representations~\cite{chan:09, chu:05}  and deep non-linear dynamics~\cite{yan:14, xie:17} were then introduced to improve the expressive power of these models. Our work took inspiration from
deep non-linear auto-regressive models, in a similar fashion to~\cite{yan:14}. However, in contrast to those cited approaches, our model can take advantage, but is not limited to, dynamic textures patterns.

\paragraph{Video frame prediction}
Recent years have sparked huge interest in conditional video prediction~\cite{fin:16, lot:17, vil:17a, vil:17b, mat:16, xue:16, shri:15, den:17, den:18, tul:18, von:17, bha:17, wich:18, gui:18, lu:17, bai:18, lee:18, xu:18, yu:18}. The goal is to generate future frames given a few frames history --a `context'. 
Most closely related to our work, some approaches  represent motion using a set of input-dependent convolution filters, that operate on an image pyramid or at image full resolution~\cite{xue:16, fin:16, bra:16, von:17}. 
Amongst those, \cite{von:17} is the only one of our knowledge which proposes a predictive model for domain-agnostic immediate future video frame generation. The authors use a sole adversial loss to constrain the network, thus accounting for the uncertainty  of the future. It is however restricted to short-term prediction, while we aim at exploring long-range solutions.  Besides, a large body of work  address the issue of disentangling video content from motion, with some applications to video synthesis~\cite{lot:17, vil:17a, den:17, hsi:18}. Most of those share a same basic principle: a dedicated network architecture, which hard-codes the decomposition between motion/pose and content, by the means of two distinct encoders, and the use of LSTM. In contrast, we propose a soft mechanism which simply learns how to distinguish the moving foreground from the background. This enables us, in particular, to recover occluded   background regions, something not possible from existing techniques.

\paragraph{Meta-learning}    
has been applied recently by Gui~\etal to tackle few-shot learning  of human-body motion prediction~\cite{gui:18}.
Metalearning techniques are based on the principle that a base-network that is properly initialized~\cite{wu:18,ha:16} (or has well-suited optimisation rules~\cite{rus:18,gui:18}), as defined from a meta-learner, can be learnt --and have good generalization properties, with a few iterations of gradient descent -and a few training samples.  While serving a very different goal ---we do not aim at any sort of meta transfer,  our model is also constructed out of two nested networks, one acting upon the other, a construct which encourages our base network to quickly reach  a robust local optimum.

\section{What can be learnt from few samples?}

{\em Motion domain}. We are primarily interested in model\-ling repetitive movements (\eg harmonic patterns such as waves, or state-space like motion of semi-rigid bodies such as a person walking), in static background scenes. Those motions are  near deterministic. We learn by watching a single period of this movement and analyse the capacity of the model to understand symmetric motions (where only half a period is given at training).

{\em  Generalization}. A conditional density model learned from scratch and from few observed frames of a given video clip is likely to memorize some aspects of the training samples  --those characteristics that are common across frames. Our model is able to generalize over the motion on future frames at test time, but partly memorizes the images appearance. 

{\em Training set}. When learning a model from few samples,  noise can be a nuisance. Invisible stochastic noise associated to clipped pixels, over-saturated images, unstable lighting or reflective surfaces can challenge the model --it tries to learn a deterministic motion from those regions.  Consequently, we prefer to process high-resolution videos (\eg $256^2$ pix), rather thas low resolution/quality clips.

{\em Interpretability}. A good motion representation is probably a representation which enables a user to understand how the model operates. We will show that the construct of our model gives us the tool for an efficient analysis of its functioning.

	\bgroup
	\begin{figure}[t]
	\begin{center}
	\begin{tabular}{cc}
	 \includegraphics[width=0.8\linewidth]{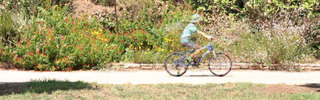}  \\
	\includegraphics[width=0.8\linewidth]{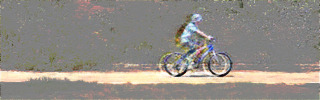} \\
    	\includegraphics[width=0.8\linewidth]{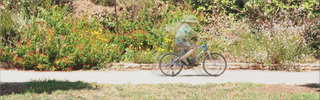} 
   	\end{tabular}
	\end{center}
	\vskip-2mm
   	\caption{The Garden sequence. Top: generated frame; middle and bottom: foreground and `background' decomposition. }
	\label{fig:fgd}
	\end{figure}
	\egroup

\section{Overview}
We aim at learning an auto-regressive sequence model $P_\zeta$, to predict $T-\delta$ future frames, given $\delta$ observed ones, $\x_{<\delta}$.  Applying the product rule, the  conditional likelihood over the future frames, $\x_{\delta:T}$, can factorized as:
	\begin{eqnarray}
	\mathcal L(\zeta) =  P_\zeta(\x_{\delta:T}|\x_{<\delta}) = \prod_{t'=\delta}^{T-1}  P_\zeta(\x_{t'+1}|\tilde \x_{t'-\delta:t'}),
	\label{eq:like}
	\end{eqnarray}
where the first frames of the time-series are observed, \ie:~$\tilde \x_{0:\delta}=\x_{0:\delta}=\x_{<\delta}$ ($\x$ refers to the ground-truth image, and $\tilde \x$ is a generated one). We use a $\delta$-order markovian assumption,  \ie  predictions are independent  conditionally of the past few frames. Future frames can be generated recursively one by one, each newly generated frame, $\tilde \x_{t}$, feeding the model for the next time step.   The set of parameters $\zeta=\{\Phi, \theta \}$ defines the model. We learn $P_\zeta$ by minimizing the negative logarithm of equation~\ref{eq:like}, so that: $\zeta = - \arg\min_\zeta \log L(\zeta)  = \arg\min_\zeta  E(\zeta)$.

 Our prediction model, $P_\zeta$, is based on two nested modules: (i) a transformation model $G_\theta$, which generates the next frame $\tilde \x_t$,  by transforming the previous ones, $\x_{t-\delta:t}$, via a series of elementary motion kernels, $W_{.,n}^l$. The size, orientation and activation amplitude of those kernels determine the transformation to be applied to the input. This encompasses both object displacement (similar to local image warping), and new pixel generation (that uncover occluded regions). (ii) a selection model, $S_\Phi$, whose role is  to choose, at each time step, which subset amongst the available motion kernels of $G_\theta$ is the most efficient to perform the desired transformation, conditioned on the  input data. Specifically, the selection model outputs a probability mass function over the kernels indices of the transformation model.  

This construct enables us  (i)  to create (\ie , learn) a generic bank of specialised (elementary) directional motion kernels; (ii)~to learn a mechanism by which a optimal   subset of kernels can be dynamically selected and applied to a given input image, in order to generate the next one. 

As a consequence, it confers to the model some key properties: (i) flexibility at test time (because the kernel selection can  adapt dynamically to each input ---as in~\cite{fin:16, bra:16}, but through a different mechanism), (ii) robustness at training time (because the net can quickly  explore very different potential solutions during the first steps of the gradient descent).

Our image transformation model can thus be written as follows:
	\begin{align} 
		   \mathcal T_\zeta:  \x_{t-\delta:t} \mapsto \tilde \x_{t+1} & = G_{\Phi, S_\Psi(t)}(  \x_{t-\delta:t} ) \\
	  & = G_\Phi( \x_{t-\delta:t} ; S_\Psi( \x_{t-\delta:t}) ). \nonumber 
	\end{align}

\section{Method}

\begin{figure}[t]
	\begin{center}
   	\includegraphics[width=0.8\linewidth]{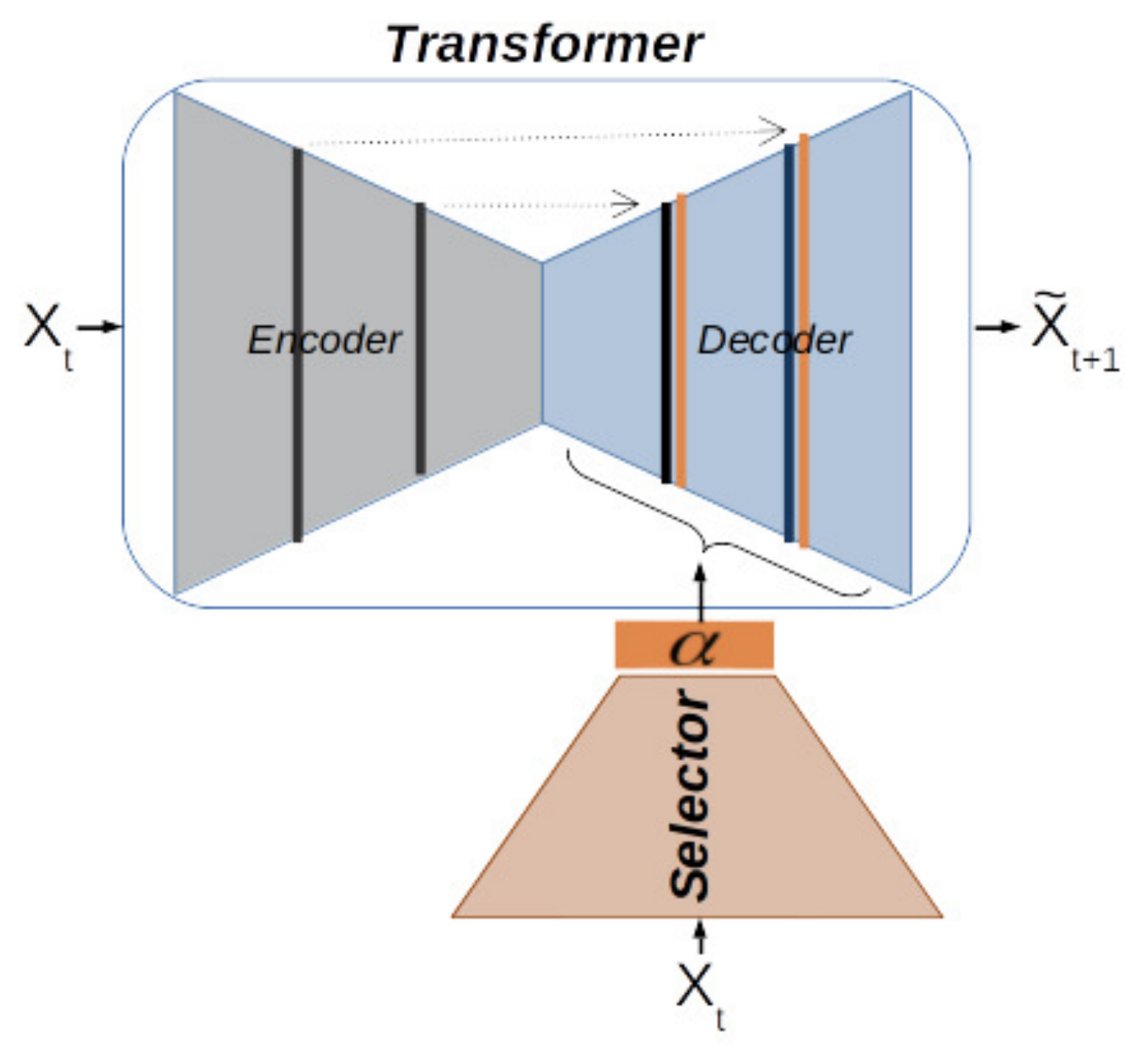}
	\end{center}
	\vskip-2mm
   	\caption{Our model consists of two nested networks: a transformer~(top) ---here represented as a 5 hidden layers encoder-decoder with skip connections, and a selector (bottom). The selector outputs weights, $\alpha(\x)$'s, that modulate the amplitude of the transformer's decoder kernels at each layer. }
	\label{fig:net}
	\end{figure}

The transformation model and selection model (or transformer and selector respectively, for short) are nested deep networks. Following~\cite{mar:17}, we advocate for a simple network architecture. The transformer, $G_\theta: \mathbb{R}^{d \times \delta} \to \mathbb{R}^{d}$, is approximated by a fully convolutional encoder-decoder network with skip connection~\cite{zhu:17}: the encoder embeds the input into a small-dimensional latent variable, while the decoder transforms this latent variable, with the help of the selector,  to generate the desired output image. The selector, $S_\Phi: \mathbb{R}^{d \times \delta} \to [0,1]^{L/2\times N} $, maps a time-dependent input onto a unit vector; $L$  and $N$ are respectively the transformer's total number of hidden layers   and the number of channels in its encoder. The models architecture is given in figure~\ref{fig:net}.

\subsection{Direction selective motion kernels}
Given an input data $\x_{\tau}$ , $\tau=[t-\delta,t]$, the selector outputs a probability  function over the transformer's elementary motion kernels indices. In practice, the selector will modify the behaviour of the transformer by modulating the amplitude of its  kernels. The selector applies only to the transformer's decoder, and ignores the encoder. 

Lets define  $ \bm{\hat \alpha}(\x_\tau)  = S_{\Phi}(\x_\tau)$  so that  $  \bm{ \hat \alpha} \in [0,1]^{L/2\times N}$ and $ \int_{n=1}^{N} \hat \alpha_n^l=1$. Then, for each building block of the {\em decoder}, the linear transformation applied to the hidden feature maps $\Y^{l-1}$ at layer  $l-1\in\{(L+1)/2,..L-1\}$,  can be defined as: 
	\begin{eqnarray}
	&	\alpha_n^l \leftarrow N \hat \alpha_n^l(\x_\tau) \nonumber \\[12pt]
	 \Z^l_{n'} &= \sum_{n=0}^{2N-1} [\Y^{L-l} ; \; \alpha^{l-1} \; \Y^{l-1}]_n * W^l_{n,n'} , 
	 \label{eq:lin}
	\end{eqnarray}
  with $n' \in \{0, ..., N-1\}$.
   $[A^a;B^b]$ refers to the concatenation of feature maps $A$ originating from the encoder at layer $a$ of the net, via the skip-connection,  with feature maps $B$ at layer $b$, along the depth/channel dimension.   In the above equation and the subsequent ones, we omit the bias term (which should write here $+b_{n}$ on the RHS), for the sake of compactness and simplicity.

Equation~\ref{eq:lin} can be developed to take  a more explicit form:
    \begin{eqnarray}
	& \Z^l_{n'} &= \sum_{n=0}^{N-1} \Y_n^{L-l} * W^l_{n,n'} + \sum_{n=N}^{2N-1}\; \Y^{l-1}_n * W^l_{n,n'} \; \alpha^{l-1}_n \nonumber \\
	&  &= N \; ( \displaystyle \mathop{\mathbb{E}}_{\Y^{L-l}_n\sim U()}[Q^{L-l}_{.,n'}] + \displaystyle \mathop{\mathbb{E}}_{\Y^{l-1}_n\sim  \alpha^l}[Q^l_{.,n'}]  \;\;\;)  \nonumber\\
	& & =  (\Z^l_{n'})^b +  (\Z^l_{n'})^f ,
	\label{eq:bf}
	\end{eqnarray}
where $*$ denote the convolution operation and $\mathop{\mathbb{E}_{\sim \mu}[M]}$ is the expected value of $M$ given its pdf $\mu$. In eq.~\ref{eq:bf}, we set $Q^{L-l}_{.,n'} = \Y_n^{L-l} * W^l_{.,n'} $, and  $Q^l_{.,n'}=\Y^{l-1}_n * W^l_{.,n'}$. 

The input and time dependant behaviour of the selected kernels is encoded in the term $ W^l_{n,n'}  \alpha_n^{l-1} = N \,  W^l_{n,n'} \hat  \alpha_n^{l-1}(\x_\tau)$ of eq.~\ref{eq:bf}, where  $ \hat \alpha^{l-1}(\x_\tau)$ is a scalar, and $W^l_{n,n'} \in \mathbb{R}^{f^2} $ are the weights of the $f\times f$ motion kernels. 

The dynamic of the transformation being encoded in the transformer's decoder exclusively, we can expect it to be in charge of modelling the motions in the scene, \ie to generate the foreground.  Conversely, the encoder might be prone to  simply learn (and remember) the scene background, that it transfers to the encoder via the skip connections.  Figure~\ref{fig:fgd} illustrates how this mechanism of foreground-background decomposition operates.

For the sake of completeness, we finally write down the expression of the very first and very last building blocks  of the transformer network:
	\begin{eqnarray}
	 \Z^0_{n'} = \sum_{t'=t-\delta}^{t}  \x_{t'} * W^0_{n,n'} , \;\;\;\;
	 \Y^0_{n'} = \rho_0(\Z^0_{n'}) , \nonumber
	\end{eqnarray}
where $\rho()$ is the non-linearity function,  and
 	\begin{eqnarray}
 	 \Z^{L}_{n'} = \sum_{n=0}^{2N-1} [\Y^{0} ; \; \alpha^{L} \; \Y^{L-1}]_n * W^L_{n,n'} , \;\;\;\;
	 \tilde \x_{t+1}  = \rho_{L}(\Z^{L}). \nonumber
	\end{eqnarray}

We  omit in this section the description of the selector, as it is similar to a classical (encoder) network. Details of the complete model architecture (number of layers, channels, non-linearity functions, \etc) are specified in the Supplementary material.

\subsection{Loss function}
We keep the loss function as simple as possible. We use the $L_1$ norm as reconstruction loss.
In addition,  we introduce a second term, a motion loss, minimizing for the total variation  in the time domain:
	\begin{eqnarray}
	&\ell_{L_1}(\x_{t}) =  |\tilde \x_t -  \x_{t}| \\
	&\ell_{motion}(\x_{t}) = \left| |\tilde \x_t - \tilde \x_{t-1}|-|\x_t - \x_{t-1}| \right|
	\end{eqnarray}

The motion loss explicitly forces  the network to account for the temporal changes between consecutive frames.
We investigate its effect in practice in the Results section.
 Hence the per-batch loss function can be written:
	\begin{eqnarray}
	E(\zeta) = \sum_{t=t'}^{t'+K} \big(\ell(\x_t)+ \mu_{motion}\mathbbm{1}_{t>t'}\ell_{motion}(\x_{t})\big) ,
	\label{eq:tloss}
	\end{eqnarray}
where $\mu_{motion}$ is a  factor weighting  the two terms, $\mathbbm{1}$ is the indicator function  and $K$ is the time-range prediction in the future at training time. Note that we do not impose any direct constraint on the output of the selector, ${\bm \alpha}(\x) = S_\Phi(\x)$.

\subsection{Training: tips and tricks}

	\begin{figure}[t]
	\begin{center}
   	\includegraphics[width=0.8\linewidth]{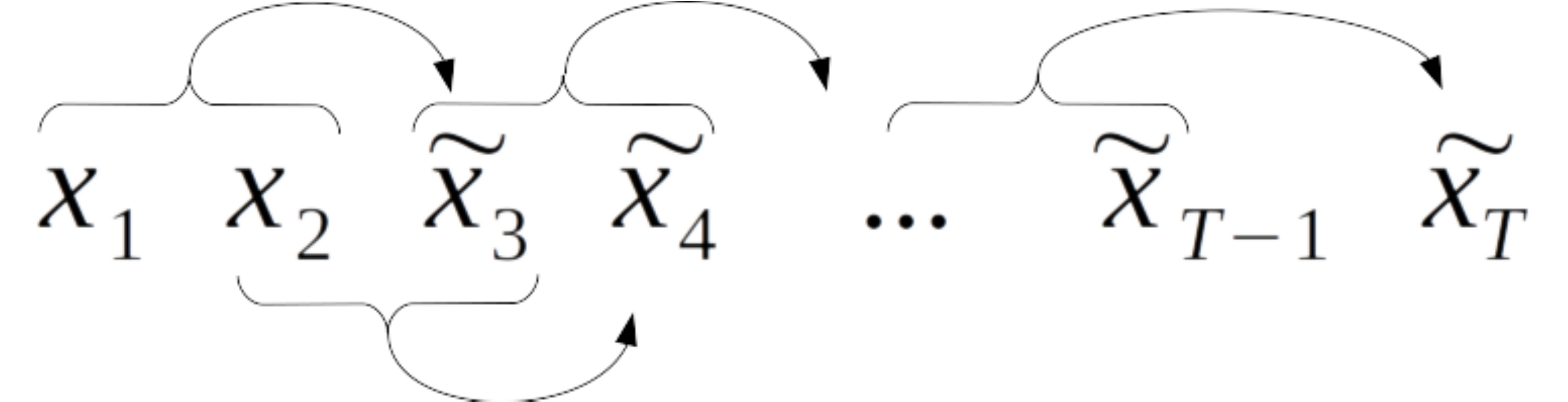}
	\end{center}
	\vskip-2mm
   	\caption{Schematic representation of the recursive prediction applied at training and testing time. $\x$ denotes ground-truth, $\tilde \x$ are predicted images. }
	\label{fig:recursive}
	\end{figure}

We train the model end-to-end in a fully unsupervised manner.  
Similar to curriculum learning (\eg~\cite{wei:18}),  we train the network with tasks of increasing difficulty. We employ training in stages:
\paragraph{Stage 1: Incremental K }
		For every new batch, we randomly pick up consecutive frames in a short temporal windows $K+\delta$ from the training set, $\x_{t'-\delta:t'+K}$, and  optimize a short form of the density, replacing $T$ by $K$ in equation~\ref{eq:like}. $K$ is increased incrementally, from $K=0$ to $K=\delta-1$, with a fixed number of batch iterations at each increment. Note that for $K=0$,  the model is conditioned on ground-truth frames only (\ie $\tilde \x = \x$), for $K=1$, a single input of the conditional set is  `fake' (\ie generated frame), and so on until $K=\delta-1$, where  a single conditioning variable is real.  We perform data augmentation by applying random left-right flips consistently among all frames of a batch.  
\paragraph{Stage 2: Full recursive learning} In this second stage, we learn to infer the whole training sequence, from the first to the last frame, following equation~\ref{eq:like} setting $T=T_{train}$, the length of the sub-sequence available for training (typically 20-30 frames). No data augmentation is performed. \\

This approach is consistent with~\cite{lu:17}.  
Besides, we use ${\bm \alpha}$  as a control variable to guide the process.  We survey the number of `active channels', \ie channels associated to a probability $\alpha_n^l>0.5$, and adjust as necessary the setting of some of the hyperparameters (learning rate and number of channels of the transformer) so that $\bm \alpha$ is not `sparse' at the end of the training.

\section{Experimental Results}

	\bgroup
	\setlength\tabcolsep{1pt}
    \def\arraystretch{0}
	\begin{figure}[t]
	\begin{center}
	\begin{tabular}{cccc}
  		\includegraphics[width=0.25\linewidth, cfbox=blue 1pt 0pt]{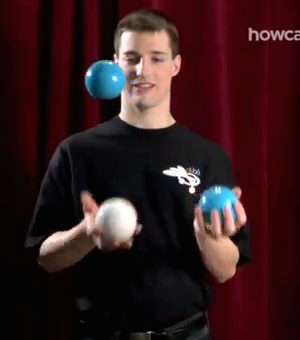} 
    		 &\includegraphics[width=0.25\linewidth, cfbox=blue 1pt 0pt]{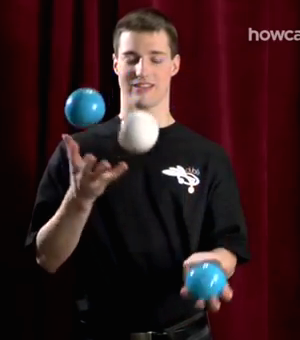}
  		 &\includegraphics[width=0.25\linewidth, cfbox=blue 1pt 0pt]{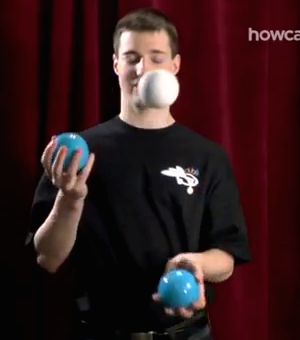} 
    		 &\includegraphics[width=0.25\linewidth, cfbox=blue 1pt 0pt]{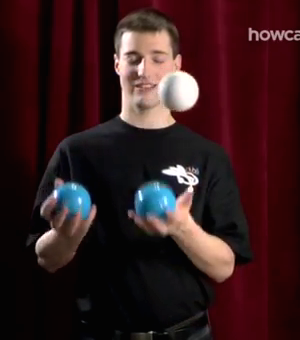}\\
  		  \includegraphics[width=0.25\linewidth, cfbox=green 1pt 0pt]{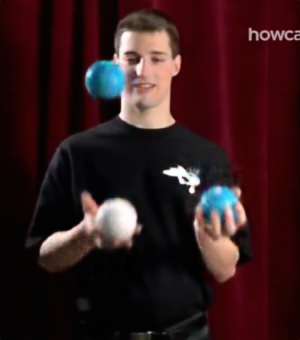} 
    		 &\includegraphics[width=0.25\linewidth, cfbox=green 1pt 0pt]{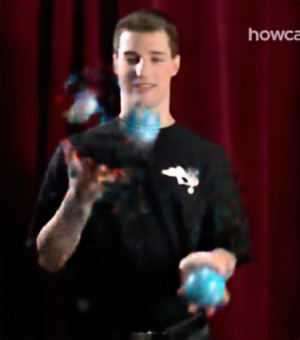} 
  		&\includegraphics[width=0.25\linewidth, cfbox=green 1pt 0pt]{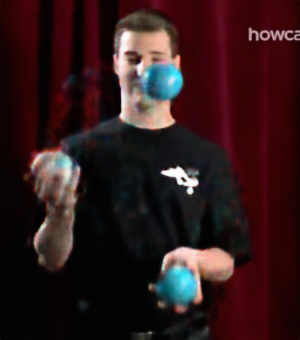} 
    		 &\includegraphics[width=0.25\linewidth, cfbox=green 1pt 0pt]{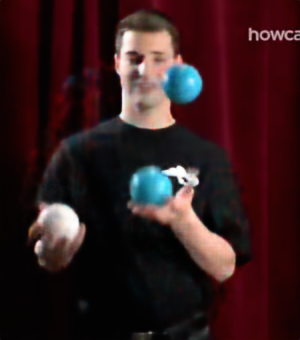}\\[3pt]
    		 \multicolumn{4}{c}{\includegraphics[width=\linewidth]{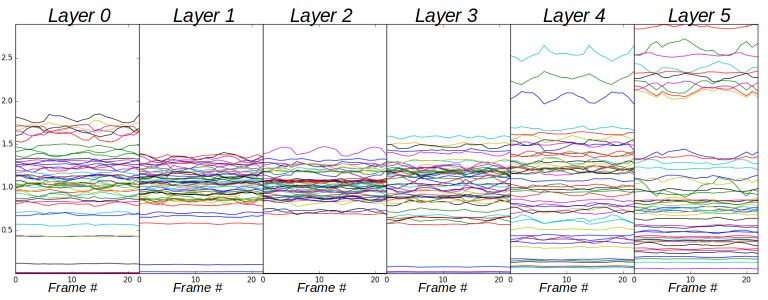}}
   	\end{tabular}
	\end{center}
	\vskip-2mm
   	\caption{Failure case due to mis-"perception" of the motion periodicity. Top: Ground-truth frames. Middle: predicted frames. Note that the color of two of the balls changes from blue to white and reciprocally. Bottom: $\alpha^l_n$ as a function of time, for $l=[0,5]$. The periodic pattern of the sequence is reflected in the $\alpha^l_n$ curves. }
	\label{fig:juggler}
	\end{figure} 
	\egroup

	\bgroup
	\setlength\tabcolsep{1pt}
    \def\arraystretch{0}
	\begin{figure}[t]
	\begin{center}
	\begin{tabular}{cc}
  		\includegraphics[width=0.45\linewidth, cfbox=blue 1pt 0pt]{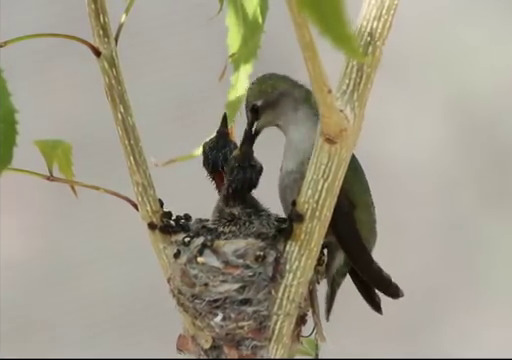} 
    		 &\includegraphics[width=0.45\linewidth, cfbox=green 1pt 0pt]{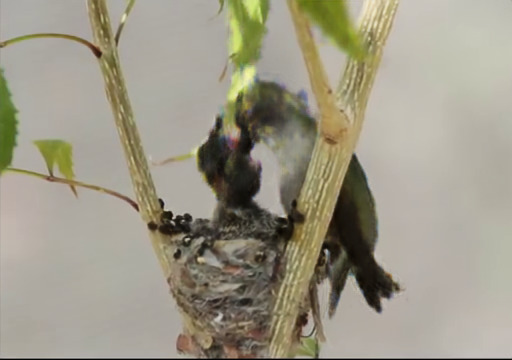}\\
  		\includegraphics[width=0.45\linewidth, cfbox=blue 1pt 0pt]{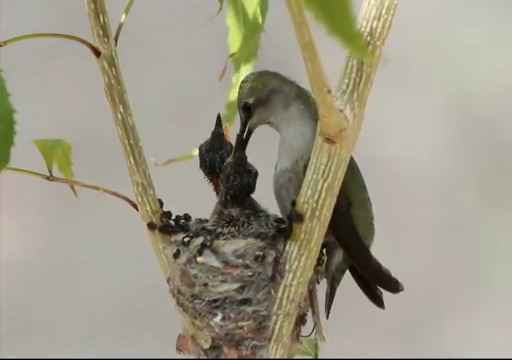} 
    		 &\includegraphics[width=0.45\linewidth, cfbox=green 1pt 0pt]{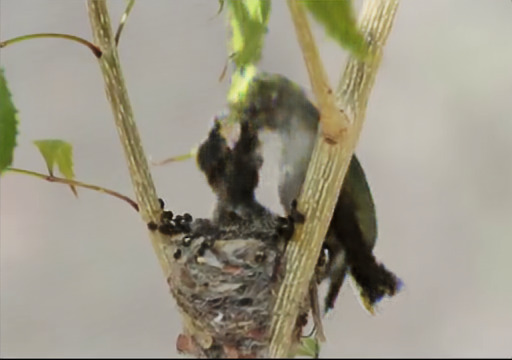}\\
  		\includegraphics[width=0.45\linewidth, cfbox=blue 1pt 0pt]{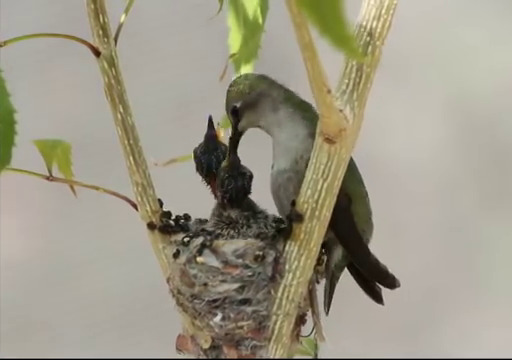} 
    		 &\includegraphics[width=0.45\linewidth, cfbox=green 1pt 0pt]{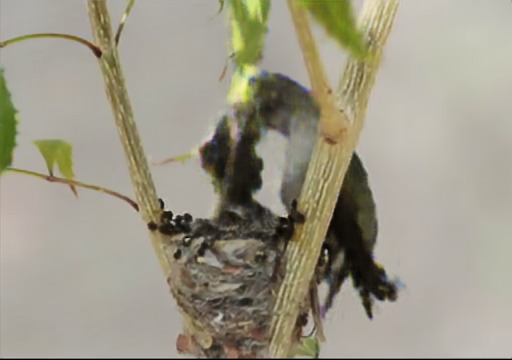} \\ [2pt]
  		\includegraphics[width=0.45\linewidth, cfbox=gray 1pt 0pt]{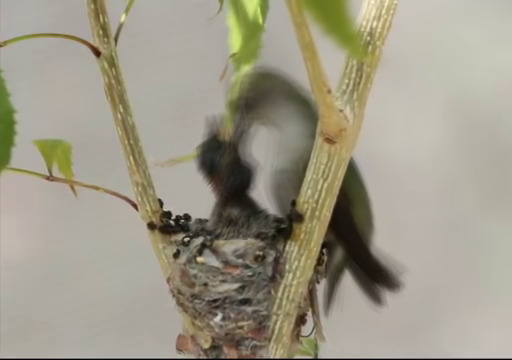} 
    		 &\includegraphics[width=0.45\linewidth, cfbox=gray 1pt 0pt]{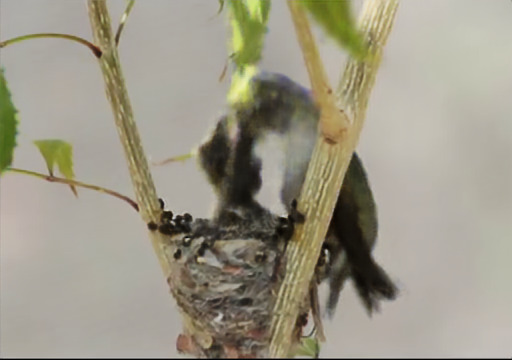}
   	\end{tabular}
	\end{center}
	\vskip-2mm
   	\caption{Failure case due to a non-deterministic motion. Top three rows: predicted (right) and associated  ground-truth images (left). Lowest raw: temporal average  of the ground-truth and predicted sequence (18 frames). The predicted frames have captured the motion of the bird, but are too blurry.}
	\label{fig:bird-failure}
	\end{figure}
	\egroup

\label{sec:res}
We evaluate our approach on several challenging sequences and attempt to analyse how the network operates on specific difficult cases. 
\subsection{Metrics and baselines} 
Following~\cite{vil:17a},  we evaluate the accuracy of the reconstruction using PSNR, SSIM~\cite{wan:04} and Mean Square measures, averaged over the length of the predicted sequence.  We run and compare: 
	\begin{enumerate} 
		\item[B0] {\em Baseline-0}. Re\-ference baseline, no prediction. We compute the error between the last input frame, and the next frame. This basic baseline informs us about the average motion amplitude (in terms of pixels change)  between two consecutive frames. 
		\item[B1] {\em Baseline-1}. Encoder-encoder. The sole transformation model $G_\theta()$ is trained, the selection model is inactive; we  set $\mu_{motion}=0$. 
		\item[M1] {\em DN w/o motion loss}. Our dual net model ---$G_\theta()$ and $S()_\Phi$ are trained jointly; we  set $\mu_{motion}=0$.
		\item[M2] {\em FDN}. Our dual net model, trained with  motion loss. We  set $\mu_{motion}=10$, unless specified otherwise.
	\end{enumerate} 

	\bgroup
	\setlength\tabcolsep{0pt}
    \def\arraystretch{0}
	\begin{figure*}[h] 
	\begin{center}
	\begin{tabular}{cccccccccc}
	\multirow{4}{*}[1.78cm]{ 
 		\includegraphics[width=0.1\linewidth,  cfbox=blue 1pt 0pt]{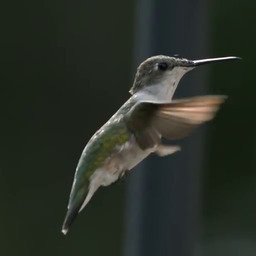} }
	& \hskip-2mm \multirow{4}{*}[1.78cm]{ 
 		\includegraphics[width=0.1\linewidth,  cfbox=blue 1pt 0pt]{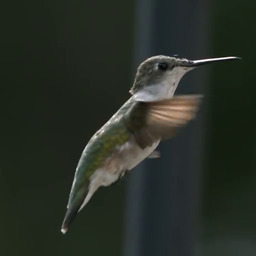} }
	& \hskip-2mm \multirow{4}{*}[1.78cm]{ 
 		\includegraphics[width=0.1\linewidth, cfbox=blue 1pt 0pt]{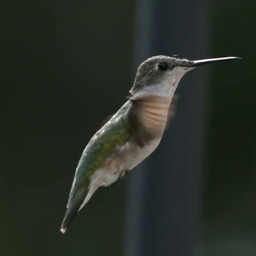} } 
    		 &\includegraphics[width=0.1\linewidth, cfbox=blue 1pt 0pt]{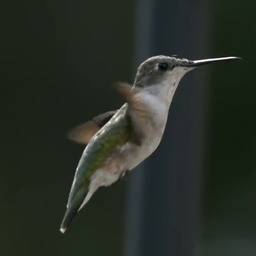}
    		 &\includegraphics[width=0.1\linewidth, cfbox=blue 1pt 0pt]{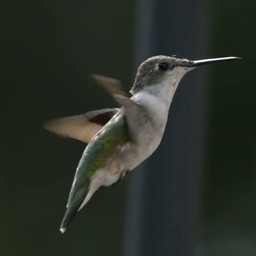}
    		 &\includegraphics[width=0.1\linewidth, cfbox=blue 1pt 0pt]{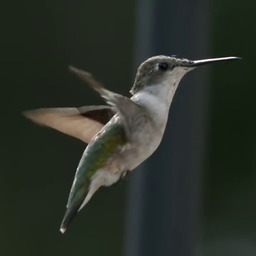}
    		  & ... 
    		 &\includegraphics[width=0.1\linewidth, cfbox=blue 1pt 0pt]{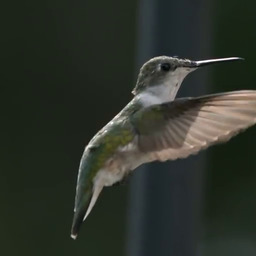}
     		&\includegraphics[width=0.1\linewidth, cfbox=blue 1pt 0pt]{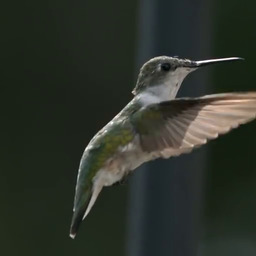}
    		 &\includegraphics[width=0.1\linewidth, cfbox=blue 1pt 0pt]{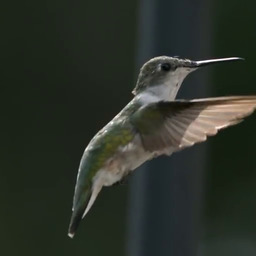} \\ 
    		 
 	& & & \includegraphics[width=0.1\linewidth, cfbox=yellow 1pt 0pt]{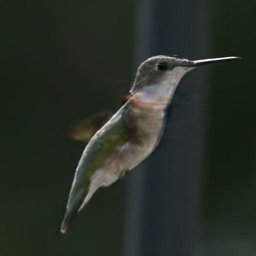}
    		 &\includegraphics[width=0.1\linewidth, cfbox=yellow 1pt 0pt]{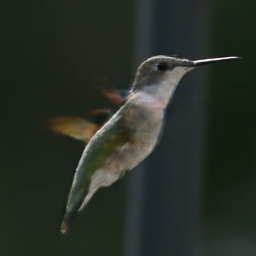}
    		 &\includegraphics[width=0.1\linewidth, cfbox=yellow 1pt 0pt]{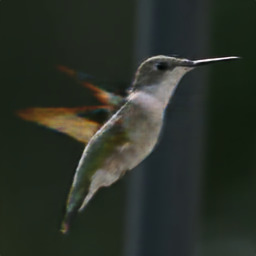}
    		  & ... 
    		 &\includegraphics[width=0.1\linewidth, cfbox=yellow 1pt 0pt]{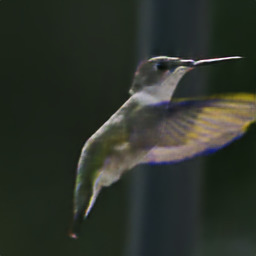}
     		&\includegraphics[width=0.1\linewidth, cfbox=yellow 1pt 0pt]{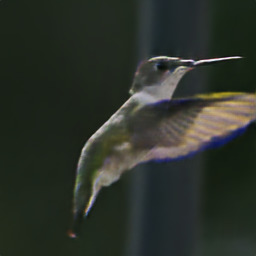}
    		 &\includegraphics[width=0.1\linewidth, cfbox=yellow 1pt 0pt]{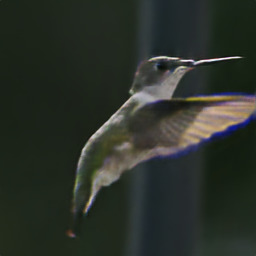} \\
    		 
   & & & \includegraphics[width=0.1\linewidth, cfbox=orange 1pt 0pt]{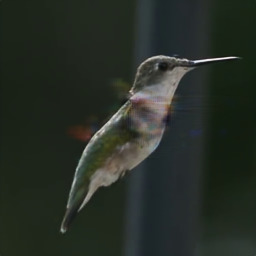}
    		 &\includegraphics[width=0.1\linewidth, cfbox=orange 1pt 0pt]{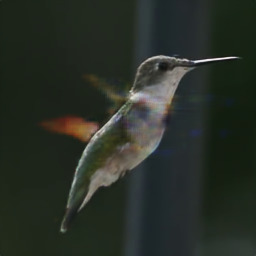}
    		 &\includegraphics[width=0.1\linewidth, cfbox=orange 1pt 0pt]{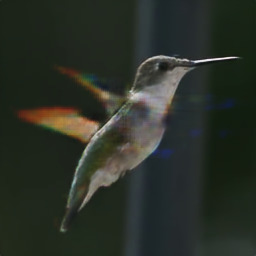}
    		  & ... 
    		 &\includegraphics[width=0.1\linewidth, cfbox=orange 1pt 0pt]{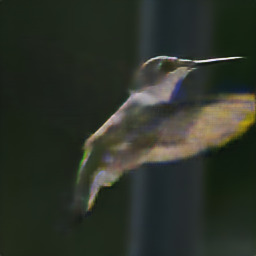}
     		&\includegraphics[width=0.1\linewidth, cfbox=orange 1pt 0pt]{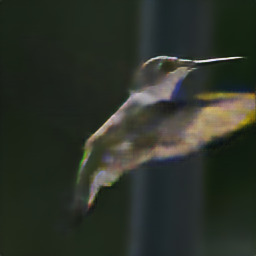}
    		 &\includegraphics[width=0.1\linewidth, cfbox=orange 1pt 0pt]{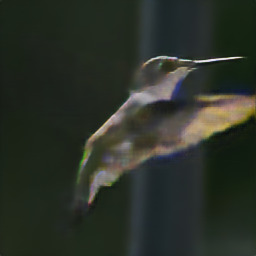} \\
    		    		     
	& & & \includegraphics[width=0.1\linewidth, cfbox=green 1pt 0pt]{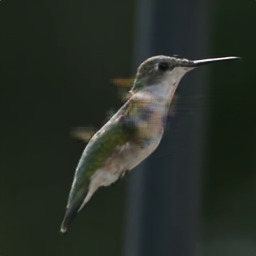}
    		 &\includegraphics[width=0.1\linewidth, cfbox=green 1pt 0pt]{shot0_predict/reconstruction_001}
    		 &\includegraphics[width=0.1\linewidth, cfbox=green 1pt 0pt]{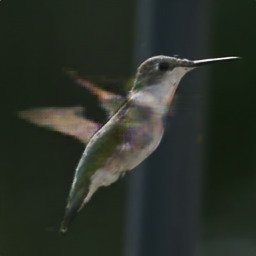}
    		  & ... 
    		 &\includegraphics[width=0.1\linewidth, cfbox=green 1pt 0pt]{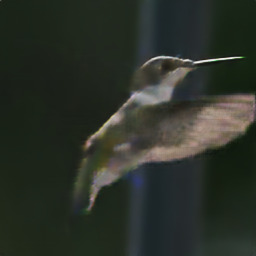}
     		&\includegraphics[width=0.1\linewidth, cfbox=green 1pt 0pt]{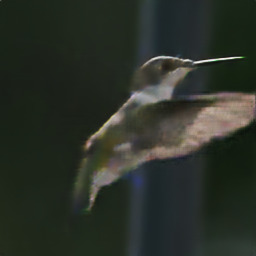}
    		 &\includegraphics[width=0.1\linewidth, cfbox=green 1pt 0pt]{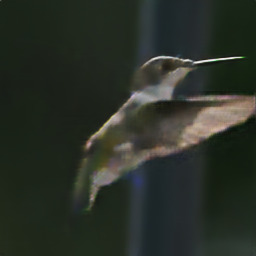} \\
    		 
    	& & & \includegraphics[width=0.1\linewidth, cfbox=gray 1pt 0pt]{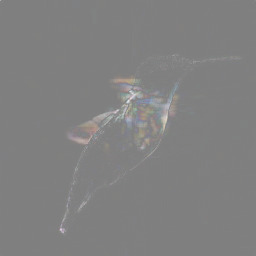}
    		 &\includegraphics[width=0.1\linewidth, cfbox=gray 1pt 0pt]{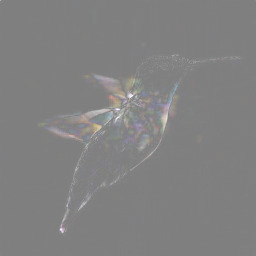}
    		 &\includegraphics[width=0.1\linewidth, cfbox=gray 1pt 0pt]{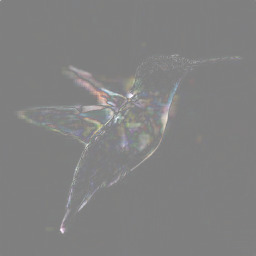}
    		  & ... 
    		 &\includegraphics[width=0.1\linewidth, cfbox=gray 1pt 0pt]{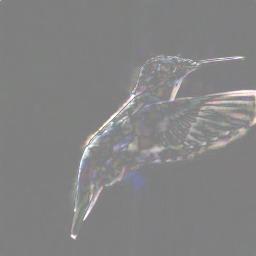}
     		&\includegraphics[width=0.1\linewidth, cfbox=gray 1pt 0pt]{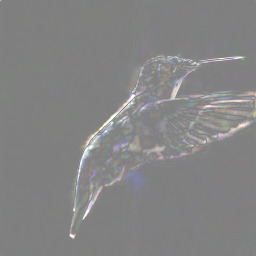}
    		 &\includegraphics[width=0.1\linewidth, cfbox=gray 1pt 0pt]{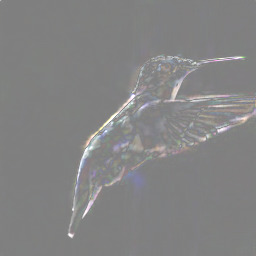} \\
    	\end{tabular}
	\end{center}
	\vskip-2mm
   	\caption{Bird sequence. Top (blue frames): input conditioning (left) and six frames (three first and three last frames, out of 25)  ground-truth (that the model has not seen at training). Yellow frames: prediction results from B1. Orange frames: M1. Green frames: M2 (our FDN). Gray frames: $L_2$ error between ground-truth and our full model (FDN) prediction. }
	\label{fig:bird}
	\end{figure*}
	\egroup

\subsection{Data and results}
{\em Bird}. The complexity of the motion of the bird flapping wings, and the subtle change of the foreground texture (semi-transparency of the wings due to the fast motion) makes the sequence challenging (figure~\ref{fig:bird}). The sequence was downloaded from Youtube, cropped and resized to $256\times 256$ pixels. It comprises 80 frames, 50 of which being used for training.  Motion is learned with a conditioning of four frames. For testing, we input to the net  four frames that it has not seen at training, and predict future 25 frames. Figure~\ref{fig:bird} shows that our approach synthesises correctly motion and appearance, while the baseline tends to introduce color artefact.

In order to test whether the network just remembers the whole sequence or adopts a smarter behaviour, we also evaluate  the prediction result obtained from a  different conditioning/entry-point (illustrations are given in the Supplementary Material). Indeed, we can show that the model `reads' correctly the given inputs and forecasts properly the next frames/poses. It suggests that, while probably the net keeps in memory some of the pose information of the bird wings, it understands the rule of transformation from one frame/pose to the next one. 

We show in figure~\ref{fig:demo} the $\alpha$ values inferred by the trained selector $S()$ as a function of time (frame \#). Each curve characterizes the variation of the amplitude of a given motion kernel at each layer of the transformer's decoder.  To generate this figure, we gave the net the first frames of the sequence, and let it predict the entire video. The last 30 frames had not been seen by the net during training. One can observe that the $\alpha$-curves reflects a periodic pattern, consistent with the bird motion. The $\alpha$-curves represent the visual motion pattern in a non-trivial way.

{\em Boy on a bicycle}. This example shows a sequence with cluttered textured background (figure~\ref{fig:garden-ocean}). The dominant motion is mainly translational,  with a rotational movement of the legs  added to it.  The sequence was acquired by a Canon EOS camera, with a resolution of $1280\time 720$, cropped then resized to $100\times 320$ pix. 
It comprises 57 frames, 30 of which being used for training. Motion is learned with a conditioning of three frames. To illustrate the results, we feed  the net with three frames of the sequence unseen at training time,  
and predict future frames until the boy leaves the camera's field of view. Results and comparison with baselines are shown in figure~\ref{fig:garden-ocean}. While all three approaches model correctly the translational motion, the  baselines either introduce some foreground color change (B1) or lose foreground details and shape contours.

In order to analyse how operated the reconstruction and foreground-background separation, we synthesise a frame by setting $(\Z^L_{n'})^b = 0$ for foreground generation, and $(\Z^L_{n'})^f = 0$  for background generation, in equation~\ref{eq:bf}. Results are illustrated in figure~\ref{fig:fgd}.  The foreground image  depicts the boy on its bicycle; it also contains a `phantom' of the bike seen in the previous frame (it appears like a shadow). 
The ground floor has been mistaken by an object in motion, probably due to its high surface reflectivity.
The $\alpha$-curves of this sequence  are constant over time, in agreement with the uniform bike's motion (see Supp. Material).

{\em Ocean}. We selected a sequence from the YUP++ dataset~\cite{der:12} depicting ocean waves and a boat moving (static camera \# 28, Ocean category), that we cropped to~$200^2$ pix and down-sampled in the time domain, to eventually get a sequence of 50 frames.  The boat displacement is uniform, while the waves are characterized by harmonic oscillations. The colors are tern, without good contrast between the boat's hull and the sea. We learn the model from 20 frames, using three frames for conditioning. We predict over the next 26 frames. Results are illustrated in fi\-gure~\ref{fig:garden-ocean}. The motion loss, accounted for only in our full model (green frames), makes here a crucial difference and allows our  model to distinguish correctly the sea from the boat's hull.\\

	\bgroup
	\setlength\tabcolsep{0pt}
    \def\arraystretch{0}
	\begin{figure*}[h]
	\begin{center}
	\begin{tabular}{ccccc}
 		\includegraphics[width=0.2\linewidth,  cfbox=blue 1pt 0pt]{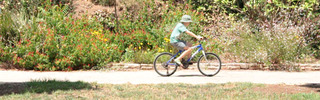}  & \\
 		\includegraphics[width=0.2\linewidth, cfbox=blue 1pt 0pt]{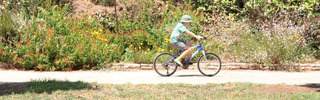}  & & \\[3pt]
 		
    		 \includegraphics[width=0.2\linewidth, cfbox=blue 1pt 0pt]{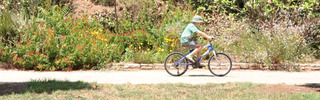} & 
    		  \includegraphics[width=0.2\linewidth, cfbox=yellow 1pt 0pt]{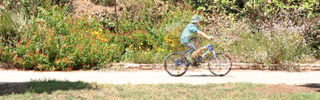} &
    		  \includegraphics[width=0.2\linewidth, cfbox=orange 1pt 0pt]{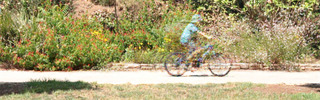} &
    		  \includegraphics[width=0.2\linewidth, cfbox=green 1pt 0pt]{garden_predict/reconstruction_000} &
    		  \includegraphics[width=0.2\linewidth, cfbox=gray 1pt 0pt]{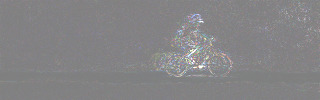} \\
    		 \includegraphics[width=0.2\linewidth, cfbox=blue 1pt 0pt]{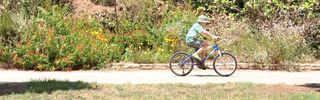} &
    		  \includegraphics[width=0.2\linewidth, cfbox=yellow 1pt 0pt]{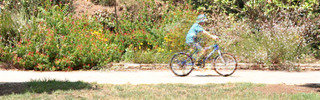} &
    		  \includegraphics[width=0.2\linewidth, cfbox=orange 1pt 0pt]{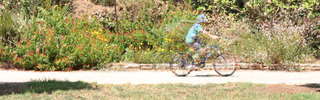} &
    		  \includegraphics[width=0.2\linewidth, cfbox=green 1pt 0pt]{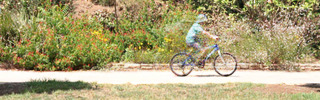} &
    		  \includegraphics[width=0.2\linewidth, cfbox=gray 1pt 0pt]{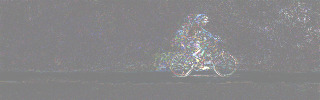} \\
    		  \includegraphics[width=0.2\linewidth, cfbox=blue 1pt 0pt]{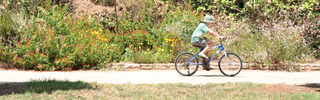} &
     	  \includegraphics[width=0.2\linewidth, cfbox=yellow 1pt 0pt]{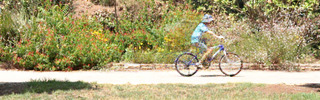} & 
    		  \includegraphics[width=0.2\linewidth, cfbox=orange 1pt 0pt]{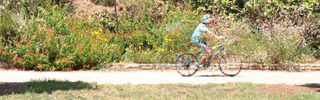} & 
   		  \includegraphics[width=0.2\linewidth, cfbox=green 1pt 0pt]{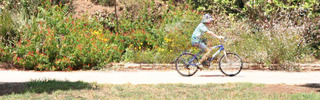} & 
    		  \includegraphics[width=0.2\linewidth, cfbox=gray 1pt 0pt]{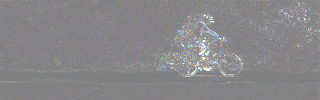} \\ [3pt]
    		  ... & ... & ... \\[3pt]
     	  \includegraphics[width=0.2\linewidth, cfbox=blue 1pt 0pt]{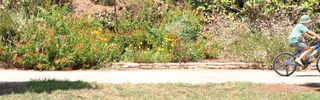} &
     	   \includegraphics[width=0.2\linewidth, cfbox=yellow 1pt 0pt]{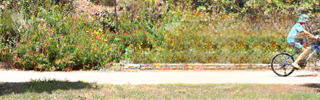} & 
     	   \includegraphics[width=0.2\linewidth, cfbox=orange 1pt 0pt]{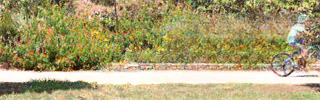} & 
     	   \includegraphics[width=0.2\linewidth, cfbox=green 1pt 0pt]{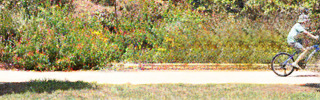} & 
     	   \includegraphics[width=0.2\linewidth, cfbox=gray 1pt 0pt]{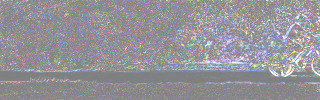}    		  	\\
     	  \includegraphics[width=0.2\linewidth, cfbox=blue 1pt 0pt]{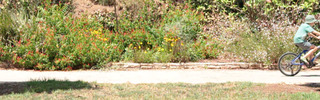} &
      	   \includegraphics[width=0.2\linewidth, cfbox=yellow 1pt 0pt]{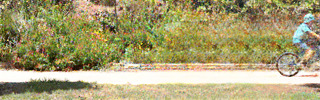} &
     	   \includegraphics[width=0.2\linewidth, cfbox=orange 1pt 0pt]{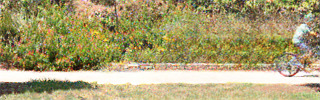} &
    	        \includegraphics[width=0.2\linewidth, cfbox=green 1pt 0pt]{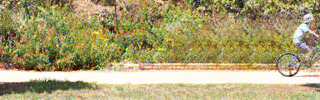} &
     	   \includegraphics[width=0.2\linewidth, cfbox=gray 1pt 0pt]{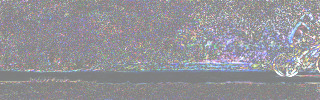} \\
    		  \includegraphics[width=0.2\linewidth, cfbox=blue 1pt 0pt]{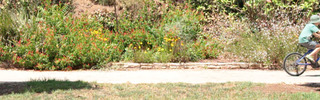} &
     	   \includegraphics[width=0.2\linewidth, cfbox=yellow 1pt 0pt]{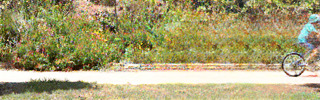} & 
     	   \includegraphics[width=0.2\linewidth, cfbox=orange 1pt 0pt]{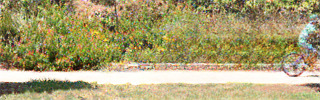} & 
   		   \includegraphics[width=0.2\linewidth, cfbox=green 1pt 0pt]{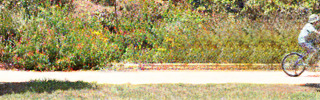} & 
    		   \includegraphics[width=0.2\linewidth, cfbox=gray 1pt 0pt]{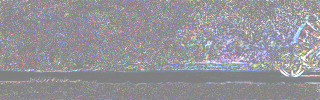} \\     
    	\end{tabular}
	\end{center}
	\vskip-2mm
   	
	\label{fig:garden}
	\end{figure*}
	\egroup
\vskip-4mm

	\bgroup
	\setlength\tabcolsep{0pt}
    \def\arraystretch{0}
	\begin{figure*}[h]
	\begin{center}
	\begin{tabular}{ccccccccc}  
	\multirow{2}{*}[2.15cm]{ 
 		\includegraphics[width=0.12\linewidth,  cfbox=blue 1pt 0pt]{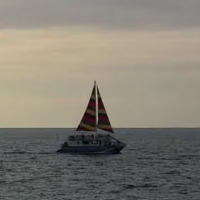} }
	& \hskip-2mm \multirow{2}{*}[2.15cm]{ 
 		\includegraphics[width=0.12\linewidth,  cfbox=blue 1pt 0pt]{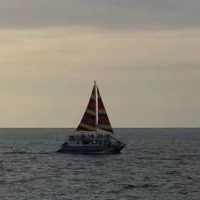} } 
    		 &\includegraphics[width=0.12\linewidth, cfbox=blue 1pt 0pt]{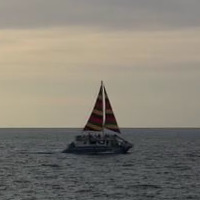}
    		 &\includegraphics[width=0.12\linewidth, cfbox=blue 1pt 0pt]{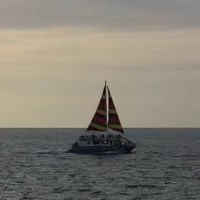}
    		 &\includegraphics[width=0.12\linewidth, cfbox=blue 1pt 0pt]{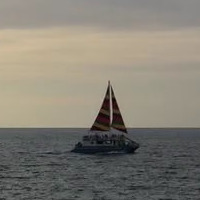}
    		  & ... 
    		 &\includegraphics[width=0.12\linewidth, cfbox=blue 1pt 0pt]{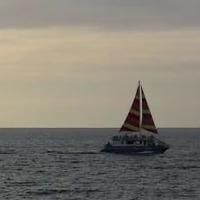}
     		&\includegraphics[width=0.12\linewidth, cfbox=blue 1pt 0pt]{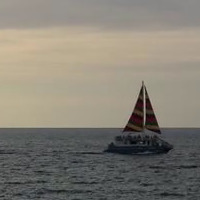}
    		 &\includegraphics[width=0.12\linewidth, cfbox=blue 1pt 0pt]{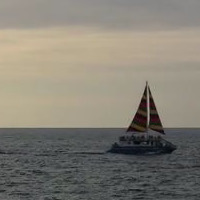} \\ 
          
	& &  \includegraphics[width=0.12\linewidth, cfbox=yellow 1pt 0pt]{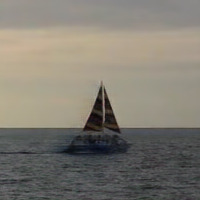}
    		 &\includegraphics[width=0.12\linewidth, cfbox=yellow 1pt 0pt]{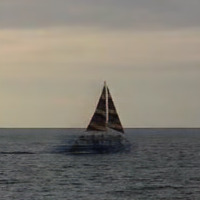}
    		 &\includegraphics[width=0.12\linewidth, cfbox=yellow 1pt 0pt]{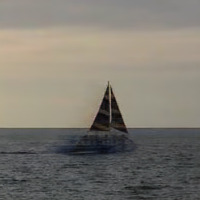}
    		  & ... 
    		 &\includegraphics[width=0.12\linewidth, cfbox=yellow 1pt 0pt]{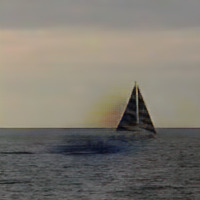}
     		&\includegraphics[width=0.12\linewidth, cfbox=yellow 1pt 0pt]{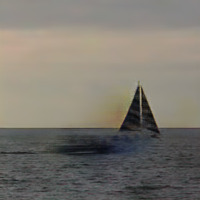}
    		 &\includegraphics[width=0.12\linewidth, cfbox=yellow 1pt 0pt]{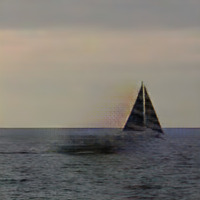} \\
    		  		     
	& &  \includegraphics[width=0.12\linewidth, cfbox=orange 1pt 0pt]{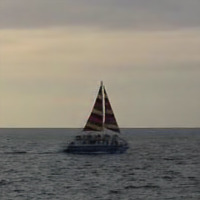}
    		 &\includegraphics[width=0.12\linewidth, cfbox=orange 1pt 0pt]{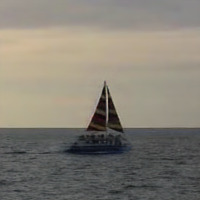}
    		 &\includegraphics[width=0.12\linewidth, cfbox=orange 1pt 0pt]{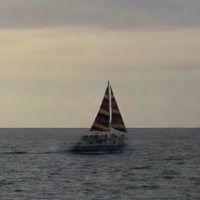}
    		  & ... 
    		 &\includegraphics[width=0.12\linewidth, cfbox=orange 1pt 0pt]{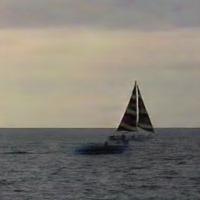}
     		&\includegraphics[width=0.12\linewidth, cfbox=orange 1pt 0pt]{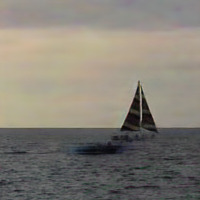}
    		 &\includegraphics[width=0.12\linewidth, cfbox=orange 1pt 0pt]{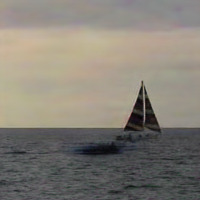} \\
    		     		 
	& &  \includegraphics[width=0.12\linewidth, cfbox=green 1pt 0pt]{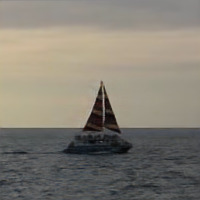}
    		 &\includegraphics[width=0.12\linewidth, cfbox=green 1pt 0pt]{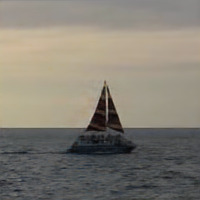}
    		 &\includegraphics[width=0.12\linewidth, cfbox=green 1pt 0pt]{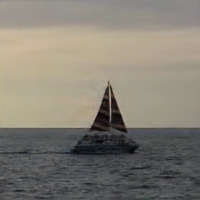}
    		  & ... 
    		 &\includegraphics[width=0.12\linewidth, cfbox=green 1pt 0pt]{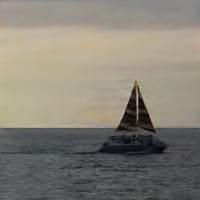}
     		&\includegraphics[width=0.12\linewidth, cfbox=green 1pt 0pt]{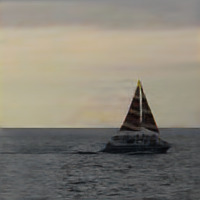}
    		 &\includegraphics[width=0.12\linewidth, cfbox=green 1pt 0pt]{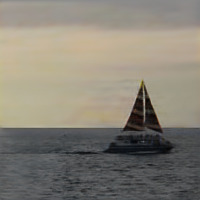} \\
    		 
	& &  \includegraphics[width=0.12\linewidth, cfbox=gray 1pt 0pt]{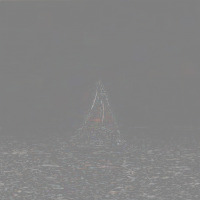}
    		 &\includegraphics[width=0.12\linewidth, cfbox=gray 1pt 0pt]{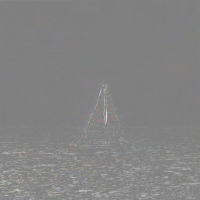}
    		 &\includegraphics[width=0.12\linewidth, cfbox=gray 1pt 0pt]{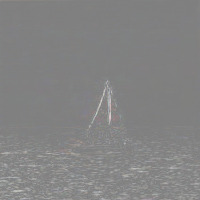}
    		  & ... 
    		 &\includegraphics[width=0.12\linewidth, cfbox=gray 1pt 0pt]{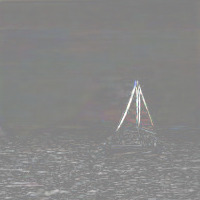}
     		&\includegraphics[width=0.12\linewidth, cfbox=gray 1pt 0pt]{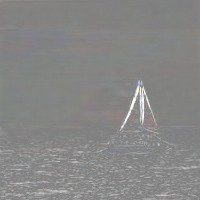}
    		 &\includegraphics[width=0.12\linewidth, cfbox=gray 1pt 0pt]{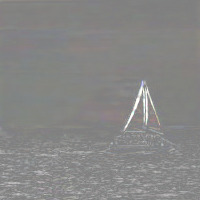} \\
    	\end{tabular}
	\end{center}
	\vskip-2mm
   	\caption{Garden (top) and Ocean (bottom) sequences. Conditioning and ground-truth (blue frames), predictions results from Baseline-1 (yellow), our DN w/o motion loss (M1, orange), and our FDN (M2, green) and associated $L_2$ error (gray). (see caption of Fig.~\ref{fig:bird})}
	\label{fig:garden-ocean}
	\end{figure*}
	\egroup

	\begin{table}
		\begin{tabular}{p{0.8cm}|c|c|c|c}
		 & B0  & B1  & M1 & M2 (FDN)\\
		\hline
		\hline
		Bird  & 22.2 &  23.1/0.913  & 23.6/0.922 & {\bf 24.2}/{\bf  0.923}\\ 
		\hline
		Garden & 19.5 & 20.3/0.682 & \bf{20.5/0.70}  & 20.42/0.695\\
		\hline
		Ocean  & 25.6 & 26.1/0.943  & 27.06/{\bf 0.963} & {\bf 27.7} /0.955\\
		\hline 
		\end{tabular}
		\caption{Quantitative analysis (average PSNR/SSIM over the predicted sequence length), for the Bird, Garden and Ocean clips. }
		\label{tab:qua}
	\end{table}

\subsection{Failure cases}
We illustrate briefly two failure cases, the Juggler (fig.~\ref{fig:juggler}) and the Bird\#2 (fig.~\ref{fig:bird-failure}). In the first case, the model understands the motion but mis-interprets its periodicity: it does not differentiate between the blue and white balls, as it should.  The $\alpha$-curves of this sequence suggests that the model builds an internal representation of the video which is invariant to the balls' color. 

The Bird\#2 case shows a limitation of the model: it does not account for the uncertainty of the future ---the motion of the bird feeding its nestling is repetitive but with high variance.

\section{Conclusion}
We have introduced a model for future frame synthesis from a single video-clip in-the-wild. Inspired initially by the mechanism of Direction Selective cells in the retina, our motion representation is based on a dual network: one that learns kernels, and a second one which dynamically selects the best subset for  next frame prediction. Our frame generations compare favourably with baseline approaches on challenging videos. 
As future work, we plan to investigate the potential of such a dual-net construct on other tasks, \eg motion composition, or motion transfer.  An other  direction  would be to extract a richer latent motion representation.

\section*{Acknowlegment}
We would like to acknowledge Michael Werman and Shmuel Peleg for helpful discussions, constructive feedback and suggestions during this project.

{\small
\bibliographystyle{ieee}
\bibliography{dirsel_cvpr19}
}
\vfill
\pagebreak
.\vfill

\pagebreak

\section*{Appendices}
We provide additional information regarding:
\begin{enumerate}
	\item Specific architecture and hyper-parameters of the model;
	\item Detailed quantitative results of the {\em Bird, Garden} and {\em Ocean} clips, that supplement the results' summary given in the main article.
	\item Additional qualitative results.
\end{enumerate}


\subsection*{A. Network architecture and hyper-parameters}
\paragraph{Preprocessing}
The input frames are normalised (\ie scaled by a factor of 255) and centralized to $[-1,1]$.
 
 \paragraph{Transformation model.}
The input tensor to the transformation model is of size $[C, H, W, T]$, where $C=\{1,3\}$ is the number of color channels, $H$ and $W$ are the dimensions of the frames, and $T=\delta$ it the length of the conditioning (typically three to four frames). The output tensor is of size $[C, H, W, 1]$. 

The net architecture is  based on a U-net, as defined in~\cite{zhu:17}. Each building block  of the encoder and of the decoder is defined as:
\textsc{ ReLu -- (De)Conv - InstanceNorm  }, except the first  and the last layer. The first one  comprises a sole  \textsc{ Conv}, and the last one substitutes the instance normalisation~\cite{uly:07} with a \textsc{tanh()} non linearity. The \textsc{ Conv } operation (resp. \textsc{ DeConv}) is performed with a stride of two (resp. upscaling of two).
Different from~\cite{zhu:17}, the number of channels (\ie the network width) of the hidden layers is constant across layers and fixed to $N$. The convolutional filters size, $f_t$, is set to $4\times 4$.

The width $N$ and the depth $L$ (\ie number of hidden layers) are set manually for each video. $N$ is set according to the complexity of the  background to model (more channels for more complex videos). $L$ is set so that the hidden feature layer of smallest spatial dimension (\ie at layer $L/2$) has a dimension of at least three pixels, at most seven. 

\begin{table}[b]
	\begin{center}
	\begin{tabular}{c|cccc}
	 & $N$ & $L$ & $ndf$  & frame size\\  
	 \hline
	 Bird & 50 & 12 & 16 & 256 $\times$ 256\\  
	 Garden & 80 & 10 & 16  & 100 $\times$ 320 \\ 
	 Ocean & 50 & 12 & 16 & 200 $\times$ 200 \\  
	 Juggler & 50 & 14 & 16 & 340 $\times$ 300\\
	 Cat & 30 & 10 & 16 & 305 $\times$ 320\\
	\end{tabular}
	\end{center}
	\caption{Hyper-parameter setting.}
\end{table}

\paragraph{Selection model.}

A  first layer of the selection model transforms the input  $\x_\tau$ into  difference images: it computes the absolute temporal difference between consecutive  frames. This process allows to discard the scene static background. 

The main architecture of the selector net is again borrowed from~\cite{zhu:17}' encoder. It takes as input a tensor of size $[C, H, W, T-1]$. It has a \textsc{ ReLu - Conv - BatchNorm } structure at each building block (except the first ones which both ignore the normalisation), with a stride of two, and  a doubling number of channels from one layer to an other. The last layer is a fully connected one. The output vector is reshaped to a matrix $N \times L/2$. The columns of the matrix are normalised with a \textsc{softmax()} function (so that $\sum_n \alpha_n^l=1$).    The number of hidden layers is set to $L/2$. The channels number at the first layer, $ndf$, is a free parameter. The convolutional filters size, $f_s$, is set to $5\times 5$.

In addition, we may add, before the fully connected layer, convolutional  blocks,  that leave unchanged the spatial dimension of the hidden layers (\ie stride one), but reduce the number of channels (divided by two at each new block).  These extra blocks are meant to reduce the size of the fully connected layer, so that the two networks (selector and transformer) are  of similar capacity (same order of magnitude).

	\begin{figure*}[t]
	\begin{center}
	\begin{tabular}{ccc}
   	\includegraphics[width=0.3\linewidth]{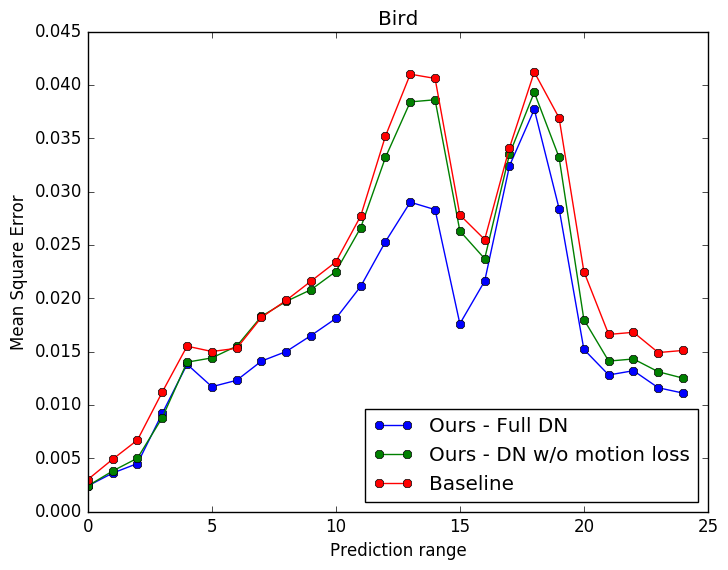}
   	\includegraphics[width=0.3\linewidth]{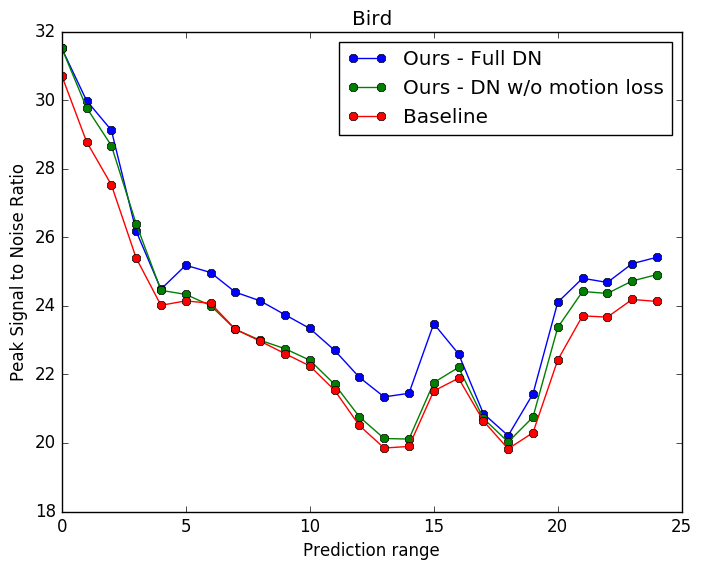}
   	\includegraphics[width=0.3\linewidth]{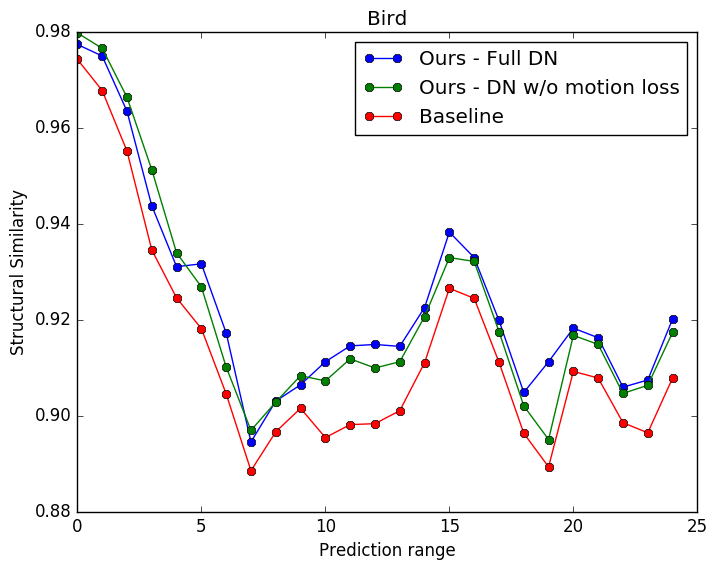}\\
   	\includegraphics[width=0.3\linewidth]{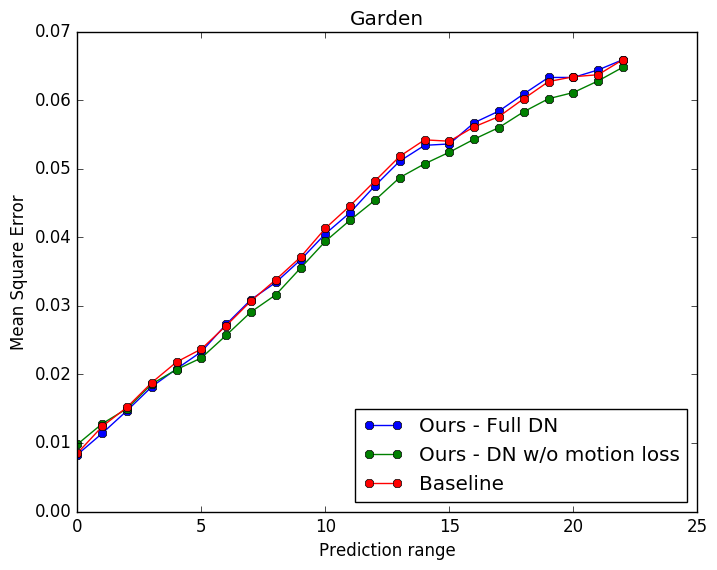}
   	\includegraphics[width=0.3\linewidth]{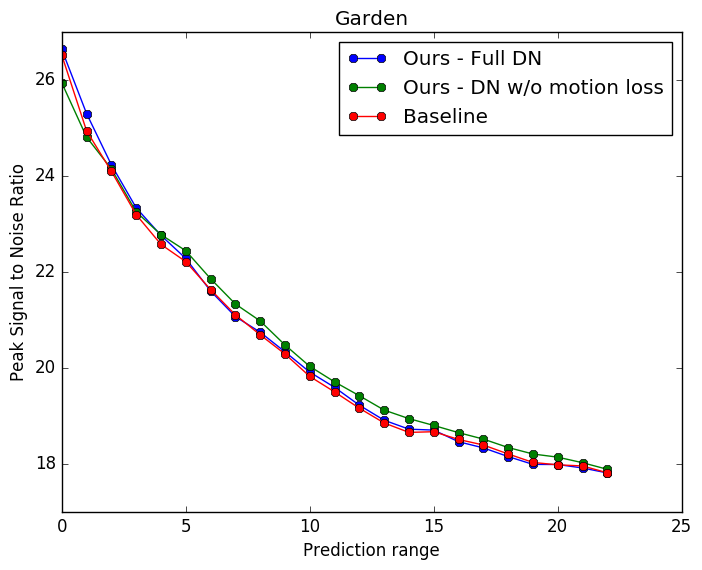}
   	\includegraphics[width=0.3\linewidth]{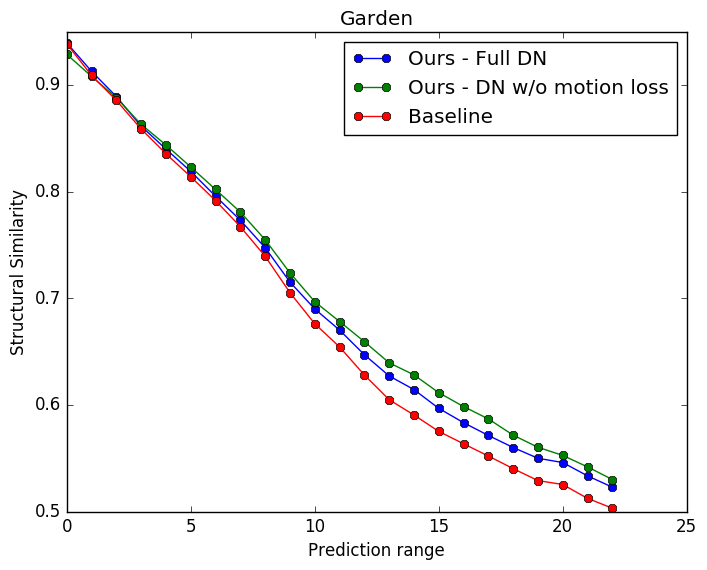}\\
   	\includegraphics[width=0.3\linewidth]{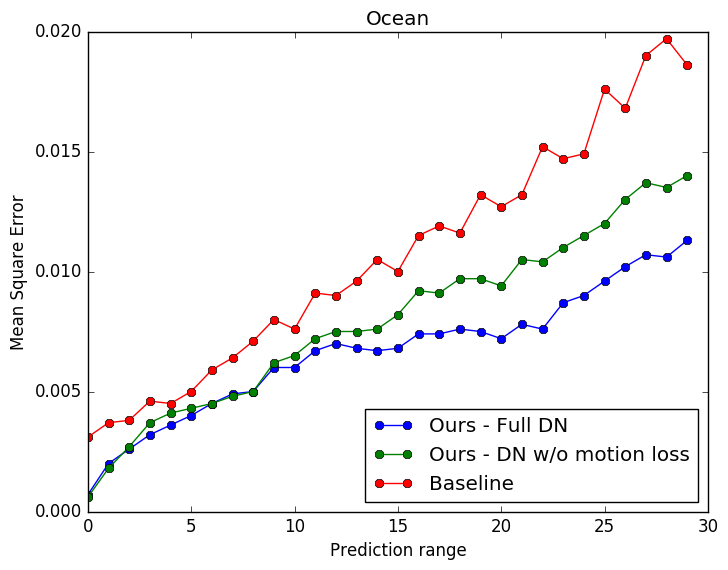}
   	\includegraphics[width=0.3\linewidth]{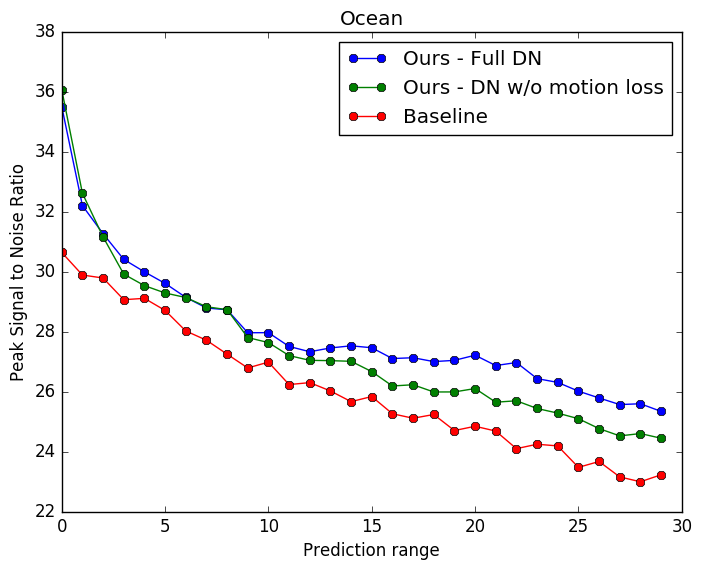}
   	\includegraphics[width=0.3\linewidth]{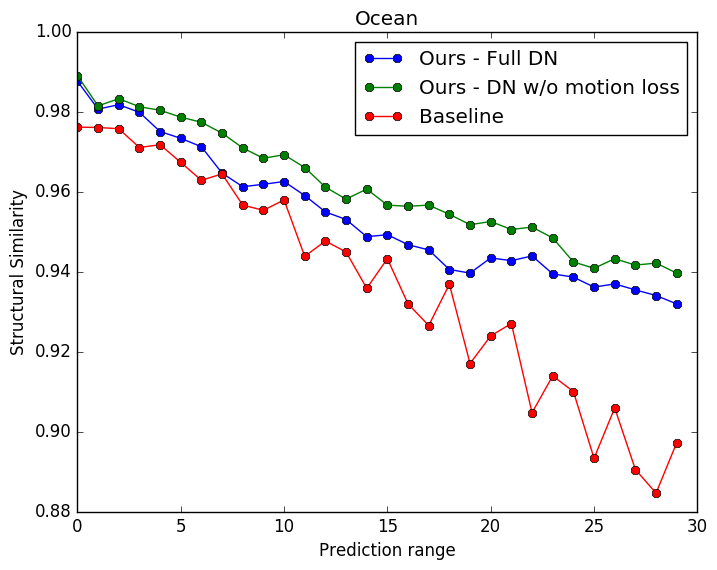}\\
   	\end{tabular}
	\end{center}
   	\caption{Quantitative comparison between our model (Dual Net) with and without motion loss and the baseline. The model is given  a few context frames, and predicts the rest of the sequence recursively, one frame at a time.}
	\label{fig:results}
	\end{figure*}

\paragraph{Learning}
We learn the model using Adam optimizer~\cite{kin:14} (beta value set to 0.9), and a initial learning rate set to either $1.e^{-3}$ or $2.e^{-4}$. The selector and the transformer weights are updated at each iteration. The learning rate decreases by a factor of two every 2K iterations. We proceed with early stopping in stage-2 of the training procedure (Section~5.3).

\subsection*{B. Detailed quantitative results}
Figure~\ref{fig:results} summarizes the quantitative comparison among: i) our Full Dual net, ii) a variation of our Dual net ---trained  with a sole $\ell_{L_1}$ loss, setting $\mu_{motion}=0$ ---, and iii) a baseline ---defined by an auto-encoder,  trained similarly with  $\mu_{motion}=0$. For our full model (full DN), the value of $\mu_{motion}$ is set to 10, except for the Ocean sequence (for which it is 1). 

We investigate the effect of the prediction range (\ie time steps into the future) on the results accuracy, using PSNR, SSIM~\cite{wan:04}, $L_2$-norm as metrics.  Our Full Dual model outperforms the two other methods, based on the MSE and PSNR metrics, on the scenes with complex  motion (\ie the Bird and the Ocean).     As expected the error increases with the time range, for the Garden and Ocean clips: the error accumulates as the foreground object (the bicycle or the boat) moves away from its original position. However, the pattern of the Bird sequence is quite different (only the wings of the bird are animated, there is no global motion). Interestingly, it suggests that the net remembers the poses of the wings (some better than other) but learn and infer the sequence of these poses.

\subsection*{C. Additional results}
\subsubsection*{Cat sequence}
The Cat sequence (figure~\ref{fig:cat}) was downloaded from Youtube, and subsampled in time and space by a factor of~2. It comprises 32 frames (105$\times$320 pixels). The motion reflects the global translational displacement  and the local movement of the cat's legs.  To illustrate the results, we feed  the net with three frames that have been seen during training (no 25 and onward), and predict over thirty frames (with no ground truth available for most of the predicted sequence). The visual comparison of our FDN results (green frames, figure\ref{fig:cat}) shows sharper contour and better motion forecast from our model, in comparison to the baselines (B1, M1).

	\bgroup
	\setlength\tabcolsep{0pt}
    \def\arraystretch{0}
	\begin{figure*}[h]
	\begin{center}
	\begin{tabular}{ccccc}
 		\includegraphics[width=0.25\linewidth,  cfbox=blue 1pt 0pt]{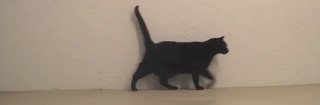}  & \\
 		\includegraphics[width=0.25\linewidth, cfbox=blue 1pt 0pt]{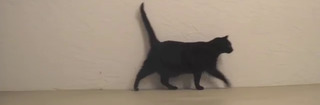}  & & \\[3pt]
 		
    		 \includegraphics[width=0.25\linewidth, cfbox=blue 1pt 0pt]{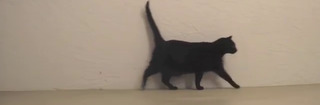} & 
    		  \includegraphics[width=0.25\linewidth, cfbox=yellow 1pt 0pt]{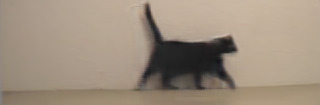} &
    		  \includegraphics[width=0.25\linewidth, cfbox=orange 1pt 0pt]{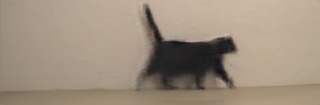} &
    		  \includegraphics[width=0.25\linewidth, cfbox=green 1pt 0pt]{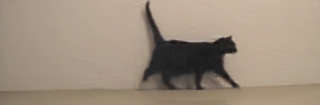}  \\
    		 \includegraphics[width=0.25\linewidth, cfbox=blue 1pt 0pt]{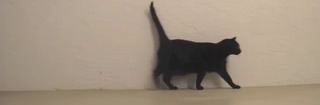} &
    		  \includegraphics[width=0.25\linewidth, cfbox=yellow 1pt 0pt]{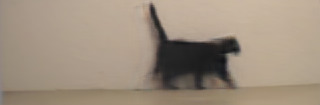} &
    		  \includegraphics[width=0.25\linewidth, cfbox=orange 1pt 0pt]{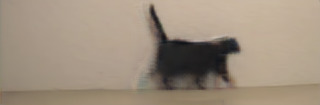} &
    		  \includegraphics[width=0.25\linewidth, cfbox=green 1pt 0pt]{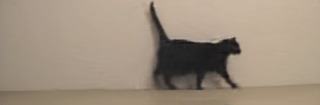}  \\
    		  \includegraphics[width=0.25\linewidth, cfbox=blue 1pt 0pt]{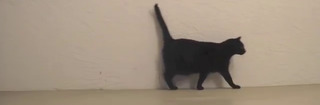} &
     	  \includegraphics[width=0.25\linewidth, cfbox=yellow 1pt 0pt]{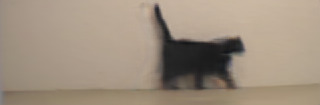} & 
    		  \includegraphics[width=0.25\linewidth, cfbox=orange 1pt 0pt]{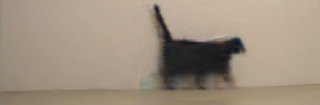} & 
   		  \includegraphics[width=0.25\linewidth, cfbox=green 1pt 0pt]{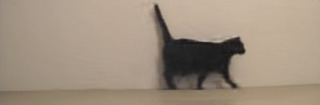}  \\ [3pt]
    		   & ... & ... \\[3pt]
     	  &
     	   \includegraphics[width=0.25\linewidth, cfbox=yellow 1pt 0pt]{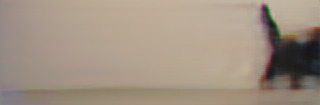} & 
     	   \includegraphics[width=0.25\linewidth, cfbox=orange 1pt 0pt]{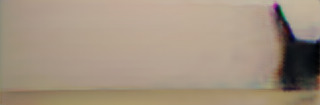} & 
     	   \includegraphics[width=0.25\linewidth, cfbox=green 1pt 0pt]{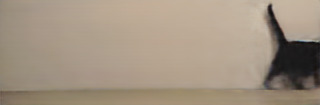}  		  	\\
     	   &
      	   \includegraphics[width=0.25\linewidth, cfbox=yellow 1pt 0pt]{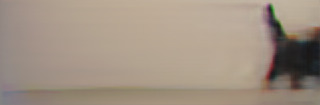} &
     	   \includegraphics[width=0.25\linewidth, cfbox=orange 1pt 0pt]{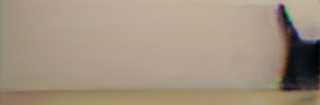} &
    	        \includegraphics[width=0.25\linewidth, cfbox=green 1pt 0pt]{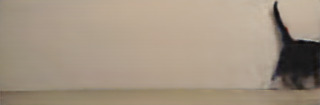}  \\
    		   &
     	   \includegraphics[width=0.25\linewidth, cfbox=yellow 1pt 0pt]{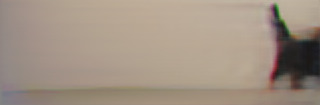} & 
     	   \includegraphics[width=0.25\linewidth, cfbox=orange 1pt 0pt]{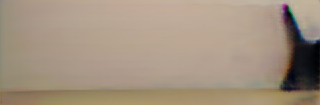} & 
   		   \includegraphics[width=0.25\linewidth, cfbox=green 1pt 0pt]{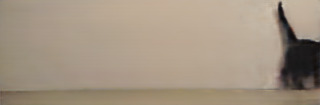}  \\     
    	\end{tabular}
	\end{center}
	\vskip-2mm
	\caption{Cat sequence. Conditioning and ground-truth (blue frames), predictions results from Baseline-1 (yellow), our DN w/o motion loss (M1, orange), and our FDN (M2, green)}
	\label{fig:cat}
	\end{figure*}
	\egroup
\vskip-4mm
 
	\bgroup
	\setlength\tabcolsep{0pt}
    \def\arraystretch{0}
	\begin{figure*}[h] 
	\begin{center}
	\begin{tabular}{ccccccccc}
	\multirow{4}{*}[1.95cm]{ 
 		\includegraphics[width=0.11\linewidth,  cfbox=blue 1pt 0pt]{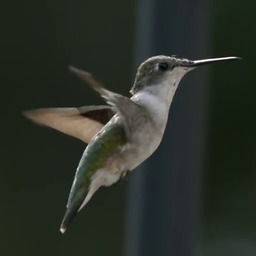} }
	& \hskip-2mm \multirow{4}{*}[1.95cm]{ 
 		\includegraphics[width=0.11\linewidth,  cfbox=blue 1pt 0pt]{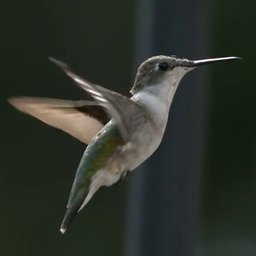} }
	& \hskip-2mm \multirow{4}{*}[1.95cm]{ 
 		\includegraphics[width=0.11\linewidth, cfbox=blue 1pt 0pt]{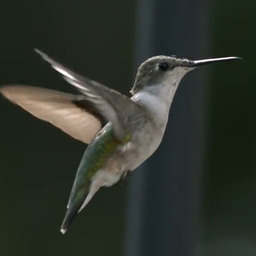} } 
    		 &\includegraphics[width=0.11\linewidth, cfbox=blue 1pt 0pt]{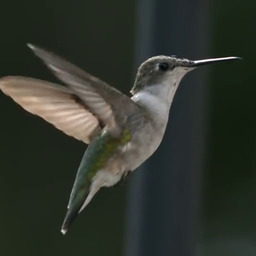}
    		 &\includegraphics[width=0.11\linewidth, cfbox=blue 1pt 0pt]{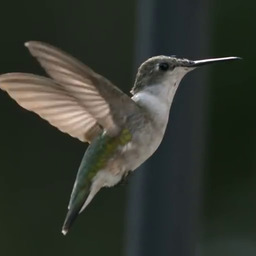}
    		 &\includegraphics[width=0.11\linewidth, cfbox=blue 1pt 0pt]{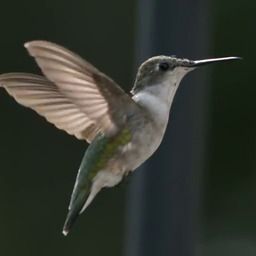}
    		 &\includegraphics[width=0.11\linewidth, cfbox=blue 1pt 0pt]{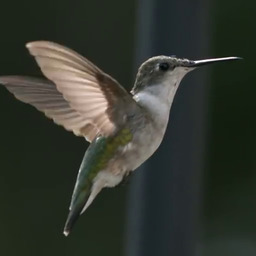}
     	 &\includegraphics[width=0.11\linewidth, cfbox=blue 1pt 0pt]{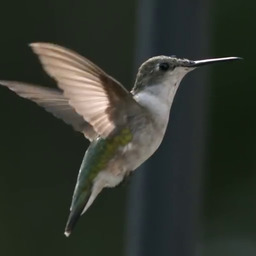}
    		 &\includegraphics[width=0.11\linewidth, cfbox=blue 1pt 0pt]{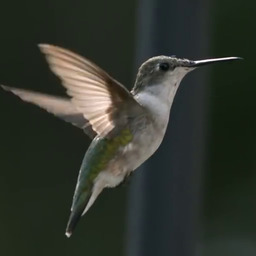} \\ 
    		    		    		     
	& & & \includegraphics[width=0.11\linewidth, cfbox=green 1pt 0pt]{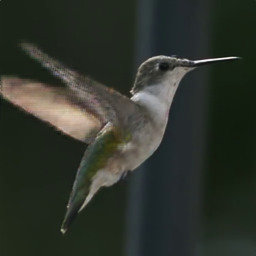}
    		 &\includegraphics[width=0.11\linewidth, cfbox=green 1pt 0pt]{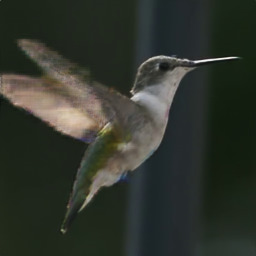}
    		 &\includegraphics[width=0.11\linewidth, cfbox=green 1pt 0pt]{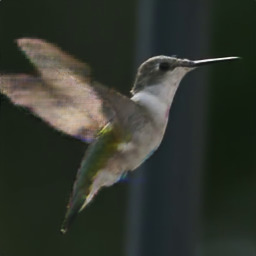}
    		 &\includegraphics[width=0.11\linewidth, cfbox=green 1pt 0pt]{shot0_predict_cond2/reconstruction_001}
     	 &\includegraphics[width=0.11\linewidth, cfbox=green 1pt 0pt]{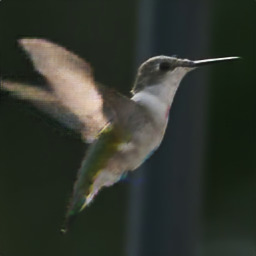}
    		 &\includegraphics[width=0.11\linewidth, cfbox=green 1pt 0pt]{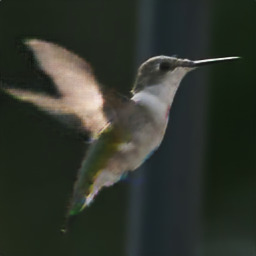} \\[5pt]
	\multirow{4}{*}[1.95cm]{ 
 		\includegraphics[width=0.11\linewidth,  cfbox=blue 1pt 0pt]{shot0_conditioning/Hummingbirds_5007} }
	& \hskip-2mm \multirow{4}{*}[1.95cm]{ 
 		\includegraphics[width=0.11\linewidth,  cfbox=blue 1pt 0pt]{shot0_conditioning/Hummingbirds_5008} }
	& \hskip-2mm \multirow{4}{*}[1.95cm]{ 
 		\includegraphics[width=0.11\linewidth, cfbox=blue 1pt 0pt]{shot0_conditioning/Hummingbirds_5009} } 
    		 &\includegraphics[width=0.11\linewidth, cfbox=blue 1pt 0pt]{shot0_gt/Hummingbirds_5010}
    		 &\includegraphics[width=0.11\linewidth, cfbox=blue 1pt 0pt]{shot0_gt/Hummingbirds_5011}
    		 &\includegraphics[width=0.11\linewidth, cfbox=blue 1pt 0pt]{shot0_gt/Hummingbirds_5012}
    		 &\includegraphics[width=0.11\linewidth, cfbox=blue 1pt 0pt]{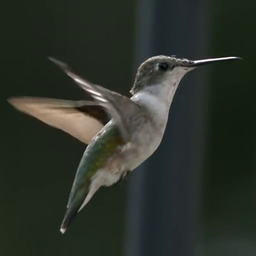}
     		&\includegraphics[width=0.11\linewidth, cfbox=blue 1pt 0pt]{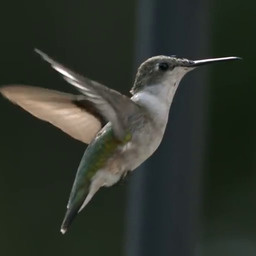}
    		 &\includegraphics[width=0.11\linewidth, cfbox=blue 1pt 0pt]{shot0_gt/Hummingbirds_5015} \\ 
    		 
	& & & \includegraphics[width=0.11\linewidth, cfbox=green 1pt 0pt]{shot0_predict/reconstruction_000}
    		 &\includegraphics[width=0.11\linewidth, cfbox=green 1pt 0pt]{shot0_predict/reconstruction_001}
    		 &\includegraphics[width=0.11\linewidth, cfbox=green 1pt 0pt]{shot0_predict/reconstruction_002}
    		 &\includegraphics[width=0.11\linewidth, cfbox=green 1pt 0pt]{shot0_predict/reconstruction_003}
     		&\includegraphics[width=0.11\linewidth, cfbox=green 1pt 0pt]{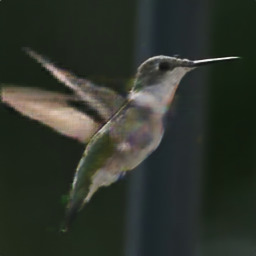}
    		 &\includegraphics[width=0.11\linewidth, cfbox=green 1pt 0pt]{shot0_predict/reconstruction_005} \\
    	\end{tabular}
	\end{center}
	\vskip-2mm
   	\caption{Bird sequence. Two different input conditionings (left, blue frames) and consecutive six frames prediction from our Full Model (Green frames). Blue frames (right) correspond to ground truth.}
	\label{fig:bird}
	\end{figure*}
	\egroup

\end{document}